\documentclass{article} 
\usepackage{iclr2027_conference,times}

\usepackage{amsmath,amsfonts,bm}

\def\eqref#1{equation~\ref{#1}}

\def\1{\bm{1}}

\DeclareMathAlphabet{\mathsfit}{\encodingdefault}{\sfdefault}{m}{sl}
\SetMathAlphabet{\mathsfit}{bold}{\encodingdefault}{\sfdefault}{bx}{n}

\usepackage{hyperref}
\usepackage{url}

\usepackage{booktabs}
\usepackage{graphicx}
\usepackage{tabularx}
\usepackage{amssymb}
\usepackage{xcolor}
\usepackage[most]{tcolorbox}
\usepackage{algorithm}
\usepackage{algpseudocode}

\usepackage[table]{xcolor}
\definecolor{OursPink}{HTML}{FFB8C4}
\definecolor{GroupGray}{HTML}{F2F2F2}
\usepackage{multirow}
\usepackage{subcaption}
\usepackage{caption}
\usepackage{titletoc}

\tcbuselibrary{breakable,skins,listings}

\definecolor{CigarBlue}{HTML}{2F5DA8}
\definecolor{CigarBlueBg}{HTML}{F3F7FF}

\newtcblisting{promptbox}[3][]{%
  breakable,
  enhanced,
  colback=#3,
  colframe=#2,
  coltitle=white,
  title={#1},
  fonttitle=\bfseries,
  boxrule=0.6pt,
  boxsep=0pt,
  arc=2pt,
  left=6pt,
  right=6pt,
  top=6pt,
  bottom=6pt,
  lefttitle=6pt,
  righttitle=6pt,
  toptitle=3pt,
  bottomtitle=3pt,
  before skip=8pt,
  after skip=8pt,
  listing only,
  listing options={%
    basicstyle=\ttfamily\footnotesize,
    breaklines=true,
    breakatwhitespace=true,
    breakindent=1.4em,
    breakautoindent=true,
    columns=fullflexible,
    keepspaces=true,
    showstringspaces=false,
    aboveskip=0pt,
    belowskip=0pt,
    literate={—}{{\textemdash}}1
  }
}

\title{Stride: Scale-Aware Multi-Agent Evolutionary Refinement for Infographic Generation}

\author{Yan Zhou \\
School of Artificial Intelligence\\
Shanghai Jiao Tong University\\
\texttt{yanzhougemzhou@sjtu.edu.cn}
\And
Yang Yang\thanks{Corresponding author.} \\
School of Artificial Intelligence\\
Shanghai Jiao Tong University\\
\texttt{angelayang@sjtu.edu.cn}
}

\iclrfinalcopy 
\begin{document}

\maketitle
\fancyhead{}
\renewcommand{\headrulewidth}{0pt}

\begin{abstract}
   Infographic generation challenges text-to-image models because a single image must satisfy numerous visual, textual, and semantic constraints. Iterative agentic refinement with a multimodal verifier is a natural remedy, but existing methods face a scale mismatch: local edits cannot correct global composition errors, while full regeneration may destroy already-correct regions. We introduce \textbf{\textsc{Stride}}, a constraint-grounded multi-agent framework that treats refinement over any frozen generator as a history-aware evolution strategy.
   \textsc{Stride} separates constraint evaluation—whether each requirement is satisfied—from visual grounding—where and how it is violated. It then routes global revisions to a composer subagent and local corrections to a retoucher subagent. Each subagent conducts internal planning and generates multiple candidates, while \textsc{Stride} accepts a candidate only when it improves the verifier score, reducing regression across iterations.
   Because stochastic generators and imperfect verifiers make any single step unreliable, it is this combination of scale-aware routing, per-step redundancy, and monotonic acceptance that yields stable improvement across iterations.
   With fully open-weight generator–verifier systems, \textsc{Stride} outperforms proprietary systems including Nano-Banana-2.0 and Seedream-5.0-Pro on BizGenEval, matches GPT-Image-2.0 and surpasses Seedream-5.0-Pro on IGenBench, and achieves a new state-of-the-art score on WISE, while substantially outperforming existing agentic methods across all evaluated benchmarks\footnote{Code is available at \url{https://github.com/SJTUSAI/Stride}.}.
\end{abstract}

\section{Introduction}

Text-to-image (T2I) generation has made remarkable progress in producing visually compelling images, but generating reliable, information-rich visual content remains challenging~\citep{li2026bizgeneval,tang2026igenbench}. Infographics are a representative example~\citep{xie2026infochartqa}, where text, numerical facts, charts, and spatial layout must work together to convey information correctly. A single incorrect value, malformed label, or inconsistent visual relationship can compromise the reliability of the entire infographic~\citep{li2026chartgalaxy,zhu2026paperbanana}. However, existing generation paradigms primarily optimize image synthesis without explicitly verifying whether generated content satisfies its underlying constraints.

Recent models have made significant progress in infographic generation. Closed-source systems such as Nano-Banana 2.0~\citep{googledeepmindNanoBanana22026} and GPT-Image 2.0~\citep{openaiIntroducingChatGPTImages} exhibit strong visual generation capabilities, while open-source models such as ERNIE-Image~\citep{liu2026ernie} and Infographic-V2~\citep{diao2026sensenova} make infographic generation increasingly accessible and specialized. Beyond direct generation, recent agent-based approaches, including PaperBanana~\citep{zhu2026paperbanana}, AutoFigure~\citep{zhu2026autofigure}, and PromptEnhancer~\citep{wang2025promptenhancer}, employ planning, feedback, and iterative refinement to further enhance generation quality. Nevertheless, existing approaches remain largely focused on visual plausibility and overall quality, with limited ability to explicitly verify fine-grained information constraints, localize their violations, and perform targeted correction.
Distilling these limitations, we identify three key challenges in current agentic generation and refinement methods:
\begin{itemize}
   \item \textbf{Insufficiently Grounded Feedback.}
         Holistic scores and coarse critiques provide limited guidance for targeted correction.
         Constraint-level judgments alone may still leave the location and nature of a failure unclear.

   \item \textbf{Mismatch between Error and Edit Scales.}
         Infographic errors range from global composition failures to localized text defects. A single revision mechanism cannot address both effectively: local edits have limited structural reach, while full-image regeneration can disrupt correct content when only a small correction is needed.

   \item \textbf{Unreliable Progress across Iterations.}
         A revision may fix one requirement while violating another, and stochastic generation makes individual proposals unreliable.
\end{itemize}

To address these challenges, we introduce \textsc{Stride}, a training-free multi-agent framework for scale-aware evolutionary refinement over frozen text-to-image generators. \textsc{Stride} separates rubric-based verification from visual grounding, combining constraint-level judgments with spatial evidence to guide targeted correction. A trajectory-aware lead editor uses this feedback and prior attempts to route revisions to a composer for global regeneration or a retoucher for localized text and formula correction. Each selected sub-agent proposes multiple candidates, and a $(1+\lambda)$-evolution strategy retains the current champion unless a candidate improves its verifier score, allowing exploration while limiting regression.
\textsc{Stride} outperforms Nano-Banana-2.0 and Seedream-5.0-Pro on BizGenEval, matches GPT-Image-2.0 and surpasses Seedream-5.0-Pro on IGenBench, and achieves a new state-of-the-art score on WISE, while substantially outperforming existing agentic methods across all evaluated benchmarks.
Our contributions are threefold:
\begin{itemize}
   \item We introduce \textsc{Stride}, a training-free framework that formulates infographic generation as scale-aware evolutionary refinement over frozen generators, adapting the scope of correction to the errors in the current image.

   \item We develop a refinement loop that separates rubric verification from spatial grounding, uses refinement history to guide global or local corrections, and combines multiple proposals with verifier-guided champion retention to reduce regression.

   \item Experiments show that \textsc{Stride} outperforms contemporary agentic methods and selected proprietary systems, while demonstrating strong scalability across different generators and benchmarks. Extensive experiments characterize the contributions of grounding, execution routes, and champion evolution.
\end{itemize}

\section{Related Works}

\textbf{Infographic Generation.}
Early text-to-image methods~\citep{wu2025qwen,diao2026sensenova} progressively improved the rendering of accurate and legible text, and later systems~\citep{gao2025postermaker,openaiIntroducingChatGPTImages} extended this to multi-layer posters and commercial visuals.
ChartGalaxy~\citep{li2026chartgalaxy} and InfoChartQA~\citep{xie2026infochartqa} study chart understanding and chart-related visual reasoning, while BizGenEval~\citep{li2026bizgeneval} and IGenBench~\citep{tang2026igenbench} evaluate whether generated visuals satisfy infographic-specified requirements.
Proprietary models~\citep{googledeepmindGemini3ProImageModelCard,googledeepmindNanoBanana22026,openaiNewChatGPTImages2025,openaiIntroducingChatGPTImages} have shown strong performance, but they still struggle to simultaneously satisfy a large number of discrete requirements and factual constraints in infographic generation.
Open-weight models~\citep{diao2026sensenova,li2026bizgeneval,tang2026igenbench} fall further behind when faced with the dense constraints imposed by infographic requirements.
\textsc{Stride} addresses this challenge through iterative dual-route evaluation and scale-aware refinement, enabling open-weight generators to match or even surpass proprietary models on this sophisticated task.

\begin{figure*}[!htbp]
   \centering
   \includegraphics[width=\textwidth]{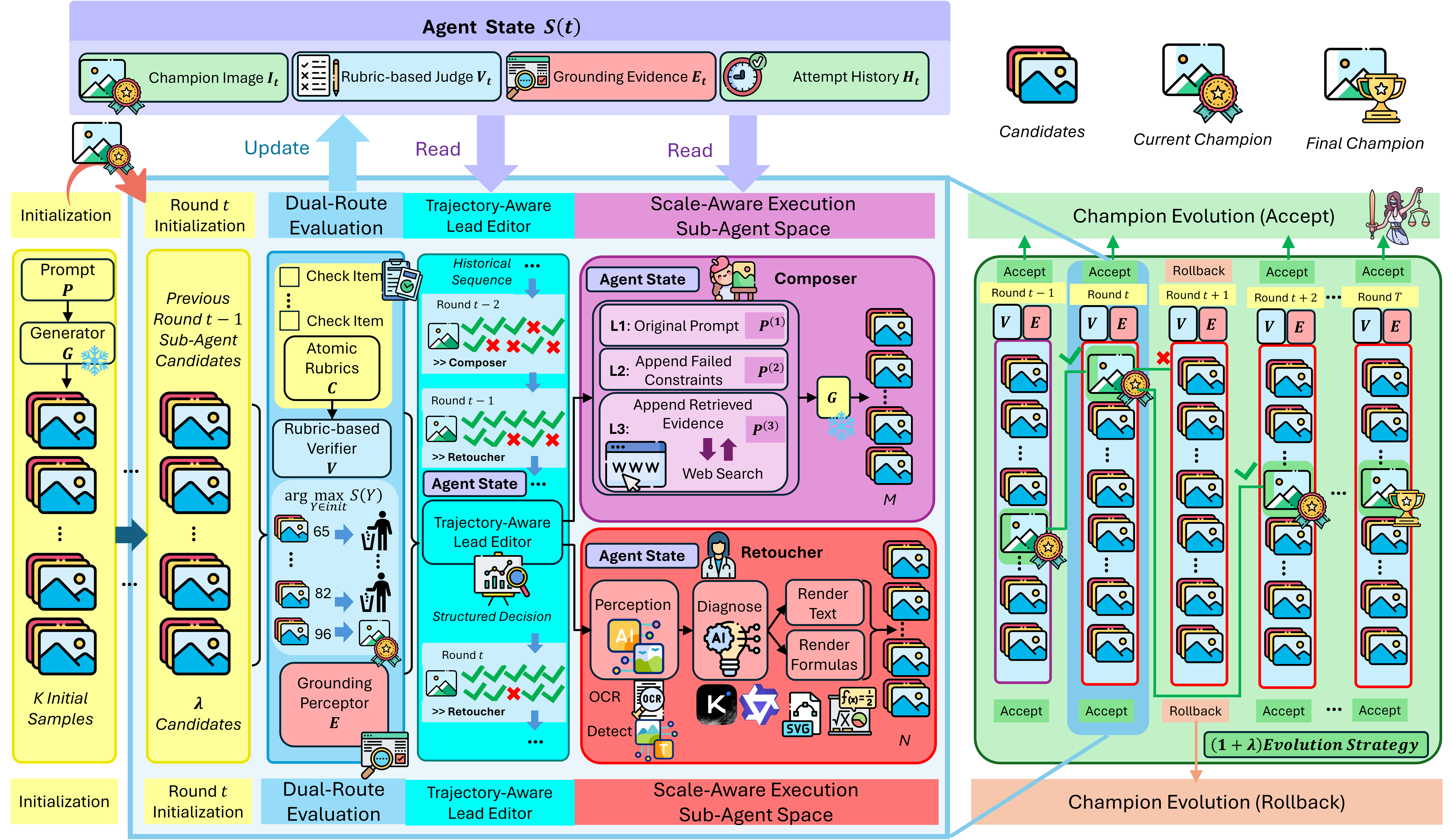}
   \caption{Overview of \textsc{Stride}.
      From $K$ initial samples, \textsc{Stride} selects a champion and refines it for $t$ rounds.
      Each round performs evolutionary refinement and agent state update through dual-route evaluation: the verifier scores candidate images against the rubrics, while the perceptor provides visual grounding of the observed failures.
      The lead editor then analyzes the agent state and refinement trajectory to route the champion to either the composer for global redraw or the retoucher for local editing.
      The generated candidates are then re-evaluated, and the champion is updated only when a candidate improves the verifier score. Repeating this process yields the final champion.}
   \label{fig:stride_pipeline}
\end{figure*}

\textbf{Agentic Augmentation.}
Agentic methods~\citep{wang2025promptenhancer,he2026mind,feng2026gen,chen2026genevolve} enhance prompts by retrieving implicit details or reference images to condition generation, but their one-shot rewriting paradigm provides limited constraint enforcement and may overwrite content that is already correct.
Methods such as Ag3~\citep{hahn2024proactive} and AutoFigure~\citep{zhu2026autofigure} use editable graphs or SVGs to specify image layouts and conduct fixed-round layout-guided refinement. However, their reliance on proprietary models obscures whether gains come from the agentic framework or the base models, which also limits compatibility with most open-weight text-to-image generators.
ReflectionFlow~\citep{zhuo2025reflection}, T2I-Copilot~\citep{chen2025t2i}, and PaperBanana~\citep{zhu2026paperbanana} use verifier feedback to assess and refine generated outputs, but the absence of explicit rubric grounding makes failures under-specified, and their reliance on prompt revision or full-image regeneration can disrupt already correct regions and limit reliable improvement.
Guided by dual-route evaluated failed rubrics and an evolution strategy, \textsc{Stride} adaptively combines global regeneration with local editing to target specific errors, addressing the above limitations while ensuring stable improvement.

\section{Method}
\label{sec:method}

\subsection{Overview}
\label{sec:overview}

As shown in Figure~\ref{fig:stride_pipeline}, \textsc{Stride} is a training-free multi-agent framework designed to iteratively refine images produced by any black-box text-to-image generator $G(P, z)$, where $P$ is a prompt query and $z$ a noise sample.
We observe that generation failures often arise in two opposite regimes: global/compositional errors that local editing cannot repair, and localized residual errors that full re-generation would only perturb.
\textsc{Stride} therefore operates over a scale-aware execution space with two complementary sub-agents -- a \textbf{composer} that globally resamples the frozen generator and a \textbf{retoucher} that locally edits the current champion.
The per-round choice between them is made by a central \textbf{lead editor}, which selects and guides the sub-agent based on the current agent state and the observations from a \textbf{dual-route evaluation} module. The core components of \textsc{Stride} are:

(1) \textbf{Dual-Route Evaluation.} \textsc{Stride} uses internal evaluations to drive iterative image refinement. Notably, it decouples evaluation into two complementary MLLM-based modules with distinct roles. The \textbf{verifier} performs closed-form judgment, scoring fine-grained semantic compliance against a rubric set. The \textbf{perceptor} performs open-form grounding, extracting region-level structured descriptions (object locations, rendered text, attribute–region bindings, etc.) that help localize where a failed rubric is violated. Separating judgment from grounding lets each module specialize, and supplies the editor with the spatial evidence a pass/fail score alone cannot provide.

(2) \textbf{Trajectory-Aware Lead Editor}. As the central controller, the lead editor maintains a running state of candidate history, prior discriminator observations and failure mode, and the detailed editing history. Rather than simple prompt rewriting, it performs reflective trajectory analysis, makes a decision on which actor (composer or retoucher) to invoke, and then formulates prompt updates for the selected actor that target to address previous failures.

(3) \textbf{Scale-Aware Execution Sub-Agent Space}. To prevent generative drift during iterative image generation, \textsc{Stride} operates over a dual-branch action space defined by two sub-agents. The \textbf{composer} explores the global generative manifold by rewriting the prompt and then resampling the image generator $G$, while the \textbf{retoucher} identifies localized editing needs and invokes specialized computer-vision tools, guarded by the perceptor's spatial grounding. The output of each editing action is then admitted or rejected under an evolution strategy. The current champion enters a new round of image refinement, until the iteration budget is exhausted.

\subsection{Agent State}
\textsc{Stride} orchestrates the image refinement process as an iterative loop. At each iteration $t$, the system maintains a state tuple $S_t$ defined as:
\begin{equation}
   S_t=\bigl(I_t,\ V_t,\ E_t,\ H_t\bigr).
   \label{eq:state}
\end{equation}
where $I_t$ denotes the current ``champion" (the highest-scoring image admitted up to iteration $t$), $V_t = V(I_t)$ represents the rubric-based judgment from the verifier, $E_t$ holds perceptor's grounding evidence (region-level localizations of where rubrics are violated) for $I_t$, and $H_t$ logs the historical sequence of executed actions, candidate rejections and failure mode, and observation deltas before step $t$. The agent state is accessible to the lead editor, the composer, and the retoucher.

\subsection{Dual-Route Evaluation}
At each iteration, candidates generated by the execution module are evaluated by a dual-discriminator measurement system. This module decouples image quality assessment into semantic rubric verification and structured perceptual evidence extraction.

\textbf{Rubric-Based Verifier.}
The rubric-based verifier is a multimodal LLM (Appendix~\ref{app:agent-prompts-verifier}) that inspects an image and answers questions according to a set of atomic natural-language rubrics $\mathcal{C} = \{c_1, c_2, \dots, c_M\}$, where each rubric specifies a fine-grained requirement of image quality. Given a candidate image $I$, the verifier outputs a binary decision $v_c(I) \in \{0, 1\}$ for each constraint $c \in \mathcal{C}$. The overall satisfaction score is defined as:
\begin{equation}
   r(I; \mathcal{C}) = \frac{1}{|\mathcal{C}|} \sum_{c \in \mathcal{C}} v_c(I).
\end{equation}
In the case of handling hierarchical or prioritized constraints, we can further partition $\mathcal{C}$ into $L$ ordered priority tiers $\big(\mathcal{C}^{(1)}, \dots, \mathcal{C}^{(L)}\big)$. An image is scored via a multi-dimensional vector:
\begin{equation}
   Score(I) = \Big(r(I; \mathcal{C}^{(1)}), \dots, r(I; \mathcal{C}^{(L)})\Big),
\end{equation}
where candidate vectors are evaluated via lexicographical ordering ($\succ$). For benchmarks with predefined checklists (e.g., BizGenEval, IGenBench), $\mathcal{C}$ is directly mapped from the evaluation checklists. For benchmarks that lack a detailed quality requirement checklist (e.g., WISE), rubrics can be dynamically constructed via a prompt parser $\mathcal{C} = \Phi(P)$, where $\Phi$ generates image-answerable constraints from the prompt (Appendix~\ref{app:constraint_interface}).

\textbf{Visual Grounding via the Perceptor.}
While the verifier issues rubric-level pass/fail judgments, the perceptor grounds those judgments spatially: it localizes objects, reads rendered text, and reports region-level attributes (Appendix~\ref{app:agent-prompts-perceptor}). This grounding resolves where and how a constraint fails, which the retoucher needs to target its local edits. The perceptor is applied to every round candidate alongside the verifier, but champion selection is driven by the verifier score $S$ alone. After each round, only the grounding of the admitted champion is written into the state as $E_t = E(I_t)$.
It's also worth noting that we tested using specialized detection/OCR tools as perceptor and they underperformed a strong MLLM model like Kimi-K3 (Appendix~\ref{app:perception_comparison}).

\subsection{Trajectory-Aware Lead Editor}
A single editing mechanism is often insufficient for iterative image refinement. When the current image contains global semantic or compositional failures, local editing is too weak to recover the missing structure. Conversely, when only a small subset of attributes remains incorrect, full-image regeneration introduces unnecessary drift. To balance these two regimes, \textsc{Stride} operates over a scale-aware action space with two complementary operators: global redraw, executed by the composer, and local refinement, executed by the retoucher.

The lead editor serves as the central controller of \textsc{Stride}. At round $t$, it
reads the current agent state  $S_t=\bigl(I_t,\ V_t,\ E_t,\ H_t\bigr)$ and performs reflective trajectory analysis. It then determines which editing sub-agent to invoke next and what editing guidance $G_t$ to pass to it (Appendix~\ref{app:agent-prompts-lead}).
The lead editor invokes the composer for global failure modes, including missing objects, incorrect layouts, and compositional errors that jointly violate multiple constraints. It invokes the retoucher when the residual errors are localized and spatially addressable, such as text-rendering errors, minor attribute mismatches, or confined visual defects.

\subsection{Scale-Aware Execution Sub-Agent Space}
\label{sec:execution}

\textbf{Global Redraw via the Composer.}
The composer produces a fresh full-image hypothesis by resampling the frozen generator under failure-aware prompt conditioning (Appendix~\ref{app:agent-prompts-composer}).
Rather than rewriting the source prompt directly, it preserves original prompt $P$ as a prefix and appends an instruction block derived from the currently failed constraints,
yielding three prompt-conditioning levels:
\begin{equation}
   P^{(\ell)}=
   \begin{cases}
      P,                                                             & \ell=1, \\[2pt]
      P \oplus A(P, \mathcal{F}_t, G_t),                             & \ell=2, \\[2pt]
      P \oplus A\bigl(P, \mathcal{F}_t, G_t, R(\mathcal{F}_t)\bigr), & \ell=3,
   \end{cases}
   \label{eq:prompt_levels}
\end{equation}
where $\oplus$ appends an instruction block while preserving $P$,
$A(\cdot)$ is produced by prompting an LLM to convert failed constraints $\mathcal{F}_t \in V_t$ into imperative guidance for the
generator, while conditioning on the lead editor's guidance $G_t$. When the composer judges that an external web search tool is needed, the retrieved evidence $R(\cdot)$ is also included in the prompt.

The composer autonomously selects between level $\ell=2$ and  $\ell=3$.
It first reasons internally from the source prompt, the failed constraints,
the verifier rationales, together with the guidance generated by the lead editor, then
emitting a candidate instruction block together with a self-assessed confidence
$\tau_c$ over the required knowledge those constraints depend on (names, dates, quantities), not over its own writing. If the confidence exceeds a threshold  $\tau_c > \tau_c^{*}$, the composer adopts the instruction block ($\ell=2$).
Otherwise, it issues self-generated queries, retrieves snippets, and revises the block against the retrieved evidence via another LLM call ($\ell=3$).
Level ($\ell=1$) is a fallback outside the model call: when no valid instruction block is produced, the framework redraws from the original prompt $P$ unchanged.
Given the
selected planning level $\ell$, the composer samples the generator $G$ to form a candidate set
\begin{equation}
   Y_t^{\mathrm{Composer}} = \{G(P^{(\ell)}, z_m)\}_{m=1}^{M},
\end{equation}
where $z_1,\dots,z_M$ are independent noise draws and $M$ is the proposal
budget for the composer.

\textbf{Local Surgery via the Retoucher.}
The retoucher improves the current champion through targeted local edits (Appendix~\ref{app:agent-prompts-retoucher}) rather
than regenerating the entire image. It first queries the perceptor for
region-level groundings $B(I_t)$: indexed OCR text blocks that localize content,
each paired with the checklist questions it fails. Because perception is
stochastic, a single pass may miss or mislocate an erroneous region. To reduce this risk, the retoucher
therefore draws $N$ parallel branches, each targeting a single region from an
independent sample of $B(I_t)$.

Branches may address the same or different regions, so parallel sampling
diversifies both localization and the proposed corrections.
In branch $n$, the retoucher reasons from the source prompt, failed
constraints, verifier rationales, lead editor's guidance and $B(I_t)$,
emitting a single edit instruction $T_n$ with confidence $\tau_r$. The instruction $T_n$ binds
one block index to a corrected string, rendered in \LaTeX{} for formulas and as plain text otherwise.
The edit is then applied to the champion, which can be formulated as:
\begin{equation}
   \tilde{I}_{t,n} = \mathrm{RenderEdit}(I_t, B(I_t), T_n).
\end{equation}

We filter branches before rendering. Branch $n$ is discarded if $\tau_r < \tau_r^{*}$ or if its edit fingerprint (i.e. the block index and its
replacement) duplicates another branch, resulting $N' \le N$ branches. Each surviving branch
is realized by the SVG toolbox, which implements $\mathrm{RenderEdit}$ locally: it
masks the selected box with the sampled background color, renders the correction
as a transparent PNG, and recovers the original typographic attributes so that the edit aligns seamlessly while the surrounding layout stays fixed.
The surviving branches form the candidate set
\begin{equation}
   Y_t^{\mathrm{Retoucher}} = \{\tilde{I}_{t,n}\}_{n=1}^{N'}, \qquad N' \le N.
\end{equation}

When an attempt is rejected, the champion is left unchanged. The grounding is
re-sampled next round, so localization keeps varying, while the lead editor's
guidance accumulates across attempts---carrying forward corrections and regression
warnings (e.g.\ an edit that repaired one block but broke a correct one) that
steer later branches away from repeated mistakes.

\subsection{Champion Evolution}
Candidates from the composer and the retoucher are placed in a shared execution space. The same measurement module scores all candidates, and the resulting verifier scores determine the champion under an evolution strategy.
At round $t$, the active sub-agent (i.e. composer or retoucher) generates a candidate set $Y_t$. We first identify the top candidate $Y_t^\star \in Y_t$ via:
\begin{equation}
   Y_t^\star = \arg\max_{Y \in Y_t} Score(Y).
\end{equation}
The champion image is updated via an \textbf{$(
      1+\lambda)$-evolution strategy} to avoid image quality drift:
\begin{equation}
   I_{t+1} =
   \begin{cases}
      Y_t^\star, & \text{if } Score(Y_t^\star) \succ Score(I_t), \\
      I_t,       & \text{otherwise}.
   \end{cases}
\end{equation}
where $1$ denotes current champion, $\lambda$ denotes the number of new champion candidates generated by the active sub-agent,  $\succ$ is the lexicographical ordering. The state $I_t$ updates strictly upon achieving a superior satisfaction vector, with ties preserving the incumbent champion. The initial champion $I_0$ is selected similarly from $K$  samples generated at $t=0$.
Each round consumes one unit of the round budget (independent of the $M$ redraw samples or $N$ retouch branches produced within it), appends unselected candidates to $H_t$, and halts upon reaching full rubric satisfaction, budget exhaustion, or execution failures. Then the lead editor starts a new round of refinement.

\section{Experiments}
\label{sec:experiments}

\subsection{Benchmarks and Setup}

\textbf{Benchmarks.}
To fairly compare \textsc{Stride} with contemporary proprietary and open-source T2I models as well as agentic generation methods, we evaluate its infographic generation and refinement capability on two large-scale public benchmarks: \textit{(1) BizGenEval}~\citep{li2026bizgeneval}, \textit{(2) IGenBench}~\citep{tang2026igenbench}.
To evaluate the scalability of \textsc{Stride}, we further conduct general-purpose, non-infographic-specific experiments on the \textit{(3) WISE}~\citep{niu2025wise} dataset.

\textbf{Setup.}
\textsc{Stride} uses Kimi-K3~\citep{team2026kimik3} as the lead editor, perceptor, composer, and retoucher, with open-weight text-to-image generators including ERNIE-Image~\citep{liu2026ernie}, Infographic-V2~\citep{diao2026sensenova}, and Qwen-Image-2512~\citep{wu2025qwen}.
The verifier is the open-weight Qwen3.5-397B-A17B~\citep{teamQwen35NativeMultimodal2026} model.
The agent runs for at most $t=3$ iterations under a $(1+\lambda)$-evolution strategy with $\lambda=4$; the initial champion $I_0$ is selected from the best-of-$K=6$ samples at $t=0$.
The composer samples $M=4$ candidates from the frozen generator, and the retoucher launches $N=4$ parallel local editing branches.
We set the internal confidence thresholds to $\tau_c^{*}=0.7$ for the composer and $\tau_r^{*}=0.8$ for the retoucher.
Final benchmark evaluation uses the corresponding official evaluator for each benchmark (Appendix~\ref{app:benchmarks}).

\subsection{Main Results}

\begin{table*}[!htbp]
   \centering
   \scriptsize
   \setlength{\tabcolsep}{2.1pt}
   \renewcommand{\arraystretch}{0.95}

   \caption{Quantitative comparison on BizGenEval.
      Each entry reports the hard score.
      The proprietary-model results are reproduced unchanged from the official BizGenEval.}

   \label{tab:bizgeneval}

   \resizebox{0.8\textwidth}{!}{%
      \begin{tabular}{@{}l | c | c | *{4}{c} | c@{}}
         \toprule
         \multirow{2}{*}{\textbf{Model}}
                                                   & \multirow{2}{*}{\textbf{Venue}}
                                                   & \multirow{2}{*}{\textbf{Params}}
                                                   & \multicolumn{4}{c|}{\textbf{Score}}
                                                   & \multicolumn{1}{c}{\textbf{Overall}}
         \\
         \cmidrule(lr){4-7}
         \cmidrule(l){8-8}
                                                   &                                      &
                                                   & \textbf{Layout}                      & \textbf{Attribute}
                                                   & \textbf{Text}
                                                   & \textbf{Knowledge}                   & \textbf{Average}
         \\
         \midrule

         \rowcolor{GroupGray}
         \multicolumn{8}{l}{\textit{Proprietary models}}
         \\

         Nano-Banana-Pro                           & 25@Google                            & --
                                                   & 72.2
                                                   & 65.6
                                                   & 86.4
                                                   & 82.6
                                                   & 76.7                                                      \\

         Wan2.6-T2I                                & 26@Alibaba                           & --
                                                   & 46.4                                 & 16.6
                                                   & 12.6
                                                   & 12.2                                 & 21.9
         \\

         Seedream-5.0                              & 26@ByteDance                         & --
                                                   & 67.6                                 & 42.4
                                                   & 43.4
                                                   & 41.8                                 & 48.8
         \\

         Seedream-5.0-Pro                          & 26@ByteDance                         & --
                                                   & 75.5                                 & 74.1
                                                   & 63.2
                                                   & 49.8                                 & 65.7
         \\

         GPT-Image-1.5                             & 26@OpenAI                            & --
                                                   & 51.6                                 & 25.8
                                                   & 40.4
                                                   & 26.0                                 & 35.9
         \\

         Nano-Banana-2.0                           & 26@Google                            & --
                                                   & 68.4                                 & 57.4
                                                   & 83.4
                                                   & 64.6                                 & 68.5
         \\

         \midrule
         \rowcolor{GroupGray}
         \multicolumn{8}{l}{\textit{Open-weight models}}
         \\

         Qwen-Image-2512                           & 25@Alibaba                           & 20B
                                                   & 22.2                                 & 1.2
                                                   & 1.8
                                                   & 0.0                                  & 6.3
         \\

         Z-Image                                   & 25@Z-Image                           & 6B
                                                   & 26.8                                 & 2.6
                                                   & 2.8
                                                   & 0.6                                  & 8.2
         \\

         HunyuanImage-3.0                          & 25@Tencent                           & 80B-A13B
                                                   & 27.8                                 & 13.8
                                                   & 10.2
                                                   & 0.0                                  & 13.0
         \\

         ERNIE-Image                               & 26@Baidu                             & 8B
                                                   & 51.8                                 & 31.8
                                                   & 28.3
                                                   & 0.1                                  & 28.0
         \\

         Infographic-V2                            & 26@SenseTime                         & 8B
                                                   & 58.8                                 & 42.6
                                                   & 63.0
                                                   & 0.0                                  & 41.1
         \\

         \midrule
         \rowcolor{GroupGray}
         \multicolumn{8}{l}{\textit{Agentic framework}}
         \\

         PE (ERNIE-Image)                          & 26@Baidu                             & 8B-3B
                                                   & 39.2                                 & 27.4
                                                   & 26.0
                                                   & 4.2                                  & 24.2
         \\

         GenEvolve (ERNIE-Image)                   & 26@Arxiv                             & 8B
                                                   & 42.8                                 & 27.0
                                                   & 18.8
                                                   & 2.8                                  & 22.9
         \\

         GenEvolve (ERNIE-Image-PE)                & 26@Arxiv                             & 8B-3B
                                                   & 35.6                                 & 20.4
                                                   & 19.4
                                                   & 6.2                                  & 20.4
         \\

         Gen-Searcher (ERNIE-Image)                & 26@Arxiv                             & 8B
                                                   & 47.6                                 & 30.2
                                                   & 23.6
                                                   & 9.4                                  & 27.7
         \\

         Gen-Searcher (Infographic-V2)             & 26@Arxiv                             & 8B
                                                   & 58.2                                 & 38.6
                                                   & \underline{58.2}
                                                   & 17.8                                 & 43.2
         \\

         PromptEnhancer-7B (ERNIE-Image)           & 26@CVPR                              & 8B
                                                   & 32.4                                 & 9.0
                                                   & 6.2
                                                   & 0.4                                  & 12.0
         \\

         PromptEnhancer-32B (ERNIE-Image)          & 26@CVPR                              & 8B
                                                   & 35.4                                 & 12.6
                                                   & 8.0
                                                   & 1.2                                  & 14.3
         \\

         PromptEnhancer-7B (Infographic-V2)        & 26@CVPR                              & 8B
                                                   & 30.6                                 & 5.8
                                                   & 2.2
                                                   & 0.6                                  & 9.8
         \\

         PromptEnhancer-32B (Infographic-V2)       & 26@CVPR                              & 8B
                                                   & 36.0                                 & 10.0
                                                   & 8.6
                                                   & 1.4                                  & 14.0
         \\

         \rowcolor{OursPink}
         \textbf{\textsc{Stride} (ERNIE-Image)}    & \textbf{Ours}                        & 8B
                                                   & \underline{64.4}                     & \underline{50.8}
                                                   & 55.2
                                                   & \underline{34.0}                     & \underline{51.1}
         \\

         \rowcolor{OursPink}
         \textbf{\textsc{Stride} (Infographic-V2)} & \textbf{Ours}                        & 8B
                                                   & \textbf{79.2}                        & \textbf{65.8}
                                                   & \textbf{79.2}
                                                   & \textbf{54.0}                        & \textbf{69.5}
         \\

         \bottomrule
      \end{tabular}%
   }
\end{table*}

\begin{table*}[t]
   \centering
   \scriptsize
   \setlength{\tabcolsep}{2.4pt}
   \renewcommand{\arraystretch}{0.95}
   \caption{Quantitative comparison on IGenBench.
      The proprietary-model results are reproduced unchanged from the official IGenBench.}
   \label{tab:igenbench_results}

   \resizebox{\textwidth}{!}{%
      \begin{tabular}{@{}l | c | c | c c c c c c c c c c | c c@{}}
         \toprule
         \multirow{2}{*}{\textbf{Model}}
                                       & \multirow{2}{*}{\textbf{Venue}}
                                       & \multirow{2}{*}{\textbf{Params}}
                                       & \multicolumn{10}{c|}{\textbf{Question Type Q-ACC}}
                                       & \multicolumn{2}{c}{\textbf{Overall}}
         \\
         \cmidrule(lr){4-13}
         \cmidrule(lr){14-15}
                                       &                                                    &                  &
         \textbf{Comp.}                & \textbf{Enc.}                                      & \textbf{Order}
                                       & \textbf{Marks}                                     & \textbf{Anno.}   &
         \textbf{Axes}
                                       & \textbf{Leg.}                                      & \textbf{Chart}   &
         \textbf{Title}                & \textbf{Deco.}
                                       & \textbf{Q-ACC}                                     & \textbf{I-ACC}
         \\
         \midrule

         \rowcolor{GroupGray}
         \multicolumn{15}{l}{\textit{Proprietary models}}
         \\

         Image-01                      & 25@MiniMax                                         & --
                                       & 0.01                                               & 0.05             & 0.04
                                       & 0.10                                               & 0.10
                                       & 0.14                                               & 0.03             & 0.22
                                       & 0.14                                               & 0.47             & 0.13             & 0.00 \\

         P-Image                       & 25@PrunaAI                                         & --
                                       & 0.08                                               & 0.15             & 0.19
                                       & 0.27                                               & 0.36
                                       & 0.28                                               & 0.54             & 0.43
                                       & 0.58                                               & 0.68             & 0.34             & 0.00 \\

         Nano-Banana                   & 25@Google                                          & --
                                       & 0.18                                               & 0.31             & 0.27
                                       & 0.44                                               & 0.54
                                       & 0.57                                               & 0.52             & 0.60
                                       & 0.65                                               & 0.81             & 0.48             & 0.02 \\

         Nano-Banana-Pro               & 25@Google                                          & --
                                       & 0.84                                               & 0.86             &
         0.90
                                       & 0.87                                               & 0.93             &
         0.93
                                       & 0.96                                               & 0.92             &
         0.98
                                       & 0.94                                               & 0.90             &
         0.49                                                                                                                            \\

         GPT-Image-1.5                 & 26@OpenAI                                          & --
                                       & 0.38                                               & 0.48             & 0.44
                                       & 0.57                                               & 0.50             & 0.54
                                       & 0.57                                               & 0.68             & 0.60
                                       & 0.80                                               & 0.55             & 0.12                    \\

         Seedream-4.5                  & 26@ByteDance                                       & --
                                       & 0.34                                               & 0.37             & 0.47
                                       & 0.48                                               & 0.70             & 0.70
                                       & 0.81                                               & 0.68             & 0.95
                                       & 0.84                                               & 0.61             & 0.06                    \\

         Seedream-5.0-Pro              & 26@ByteDance                                       & --
                                       & 0.64                                               & 0.47             & 0.71
                                       & 0.73                                               & 0.77             & 0.85
                                       & 0.83                                               & 0.83             & 0.86
                                       & 0.88                                               & 0.73             & 0.17                    \\

         GPT-Image-2.0                 & 26@OpenAI                                          & --
                                       & 0.89                                               & 0.61             & 0.93
                                       & 0.87                                               & 0.94             & 0.98
                                       & 0.95                                               & 0.93             & 0.99
                                       & 0.95                                               & 0.88             & 0.38                    \\

         Qwen-Image-3.0                & 26@Alibaba                                         & --
                                       & 0.65                                               & 0.43             & 0.77
                                       & 0.77                                               & 0.84             & 0.84
                                       & 0.87                                               & 0.85             & 0.80
                                       & 0.88                                               & 0.74             & 0.13                    \\
         \midrule
         \rowcolor{GroupGray}
         \multicolumn{15}{l}{\textit{Open-weight models}}
         \\

         Qwen-Image                    & 25@Alibaba                                         & 20B
                                       & 0.10                                               & 0.13             & 0.19
                                       & 0.29                                               & 0.43
                                       & 0.37                                               & 0.51             & 0.48
                                       & 0.56                                               & 0.78             & 0.36             & 0.01 \\

         SenseNova-U1                  & 26@SenseTime                                       & 8B-A3B
                                       & 0.17                                               & 0.22             & 0.33
                                       & 0.41                                               & 0.36
                                       & 0.49                                               & 0.56             & 0.60
                                       & 0.55                                               & 0.78             & 0.42             & 0.02 \\

         ERNIE-Image                   & 26@Baidu                                           & 8B
                                       & 0.17                                               & 0.16             & 0.25
                                       & 0.34                                               & 0.43
                                       & 0.47                                               & 0.61             & 0.55
                                       & 0.57                                               & 0.81             & 0.41             & 0.01 \\

         Infographic-V2                & 26@SenseTime                                       & 8B
                                       & 0.54                                               & 0.49             & 0.69
                                       & 0.66                                               & 0.79
                                       & 0.80                                               & 0.85             & 0.80
                                       & 0.94                                               & 0.88
                                       & 0.72                                               & 0.17
         \\

         \midrule
         \rowcolor{GroupGray}
         \multicolumn{15}{l}{\textit{Agentic framework}}
         \\

         Gen-Searcher (ERNIE-Image)    & 26@Arxiv                                           & 8B
                                       & 0.17                                               & 0.12             & 0.30
                                       & 0.39                                               & 0.44
                                       & 0.46                                               & 0.53             & 0.58
                                       & 0.50                                               & 0.77
                                       & 0.40                                               & 0.01
         \\

         Gen-Searcher (Infographic-V2) & 26@Arxiv                                           & 8B
                                       & \underline{0.39}                                   & 0.29             & \underline{0.54}
                                       & 0.51                                               & \underline{0.73}
                                       & \underline{0.69}                                   & 0.78             & 0.68
                                       & \underline{0.90}                                   & 0.84
                                       & \underline{0.60}                                   & \underline{0.07}
         \\

         PromptEnhancer-32B (ERNIE-Image)
                                       & 26@CVPR                                            & 8B
                                       & 0.21                                               & 0.24             & 0.27
                                       & 0.39                                               & 0.47
                                       & 0.47                                               & 0.70             & 0.60
                                       & 0.59                                               & 0.80
                                       & 0.45                                               & 0.01
         \\

         PromptEnhancer-32B (Infographic-V2)
                                       & 26@CVPR                                            & 8B
                                       & 0.29                                               & 0.17             & 0.32
                                       & 0.46                                               & 0.55
                                       & 0.56                                               & \underline{0.80} & 0.60
                                       & 0.89                                               & 0.80
                                       & 0.52                                               & 0.05
         \\

         \rowcolor{OursPink}
         \textbf{\textsc{Stride} (ERNIE-Image)}
                                       & \textbf{Ours}                                      & 8B
                                       & 0.30                                               & \underline{0.32} & 0.41
                                       & \underline{0.52}                                   & 0.60
                                       & 0.63                                               & 0.73             & \underline{0.70}
                                       & 0.67                                               & \underline{0.89}
                                       & 0.55                                               & 0.03
         \\

         \rowcolor{OursPink}
         \textbf{\textsc{Stride} (Infographic-V2)}
                                       & \textbf{Ours}                                      & 8B
                                       & \textbf{0.70}                                      & \textbf{0.72}    & \textbf{0.81}
                                       & \textbf{0.78}                                      & \textbf{0.90}
                                       & \textbf{0.89}                                      & \textbf{0.92}    & \textbf{0.91}
                                       & \textbf{0.98}                                      & \textbf{0.94}
                                       & \textbf{0.84}                                      & \textbf{0.38}
         \\

         \bottomrule
      \end{tabular}%
   }
\end{table*}

\textbf{BizGenEval.}
Table~\ref{tab:bizgeneval} shows that \textsc{Stride} substantially improves both open-weight generators, outperforms existing agentic baselines and proprietary systems including Nano-Banana-2.0~\citep{googledeepmindNanoBanana22026} and Seedream-5.0-Pro~\citep{bytedanceSeedream50Pro}.
With ERNIE-Image and Infographic-V2, \textsc{Stride} achieves overall Hard scores of 51.1 and 69.5, improving over the corresponding frozen generators by 23.1 and 28.4 points, respectively.
It also exceeds the strongest agentic baseline--Gen-Searcher~\citep{feng2026gen} on the two generators by 23.4 and 26.3 points.
These gains are consistent across all four capabilities: for Infographic-V2, layout, attribute, text, and knowledge scores improve by 20.4, 23.2, 16.2, and 54.0 points, respectively, indicating that \textsc{Stride} enhances factual and constraint-level correctness rather than merely improving visual layout.
Notably, using only open-weight components, \textsc{Stride} (Infographic-V2) attains a score of 69.5, surpassing the proprietary Nano-Banana-2.0 (68.5) and Seedream-5.0-Pro (65.7), demonstrating strong competitiveness.

\textbf{IGenBench.}
As presented in Table~\ref{tab:igenbench_results},
using Infographic-V2, \textsc{Stride} achieves 0.84 Q-ACC and 0.38 I-ACC, compared with 0.72 and 0.17 for the frozen generator and 0.60 and 0.07 for the strongest agentic baseline, Gen-Searcher~\citep{feng2026gen}.
\textsc{Stride} (ERNIE-Image) also improves over its base generator, increasing Q-ACC from 0.41 to 0.55.
With all open-weight components, \textsc{Stride} (Infographic-V2) attains an I-ACC of 0.38, matching GPT-Image-2.0~\citep{openaiIntroducingChatGPTImages}.
Moreover, its Q-ACC and I-ACC both surpass those of advanced proprietary systems, including Nano-Banana~\citep{googledeepmindGemini2.5FlashImageModelCard}, Seedream-5.0-Pro~\citep{bytedanceSeedream50Pro}, and Qwen-Image-3.0~\citep{teamQwen-Image-3.0}.

\begin{figure*}[t]
   \centering
   \includegraphics[width=\textwidth]{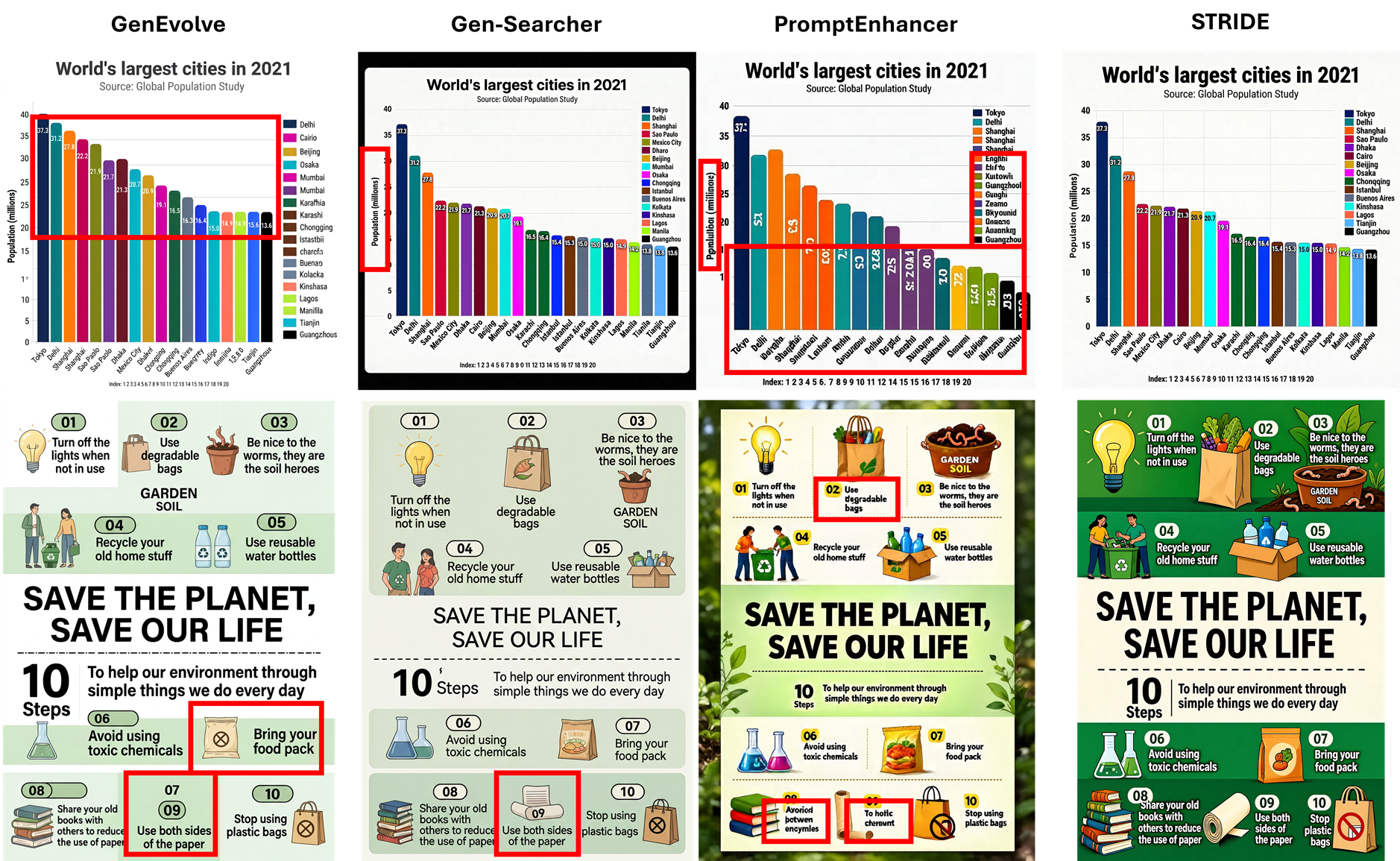}
   \caption{Qualitative comparison with existing agentic methods. Red boxes highlight errors.}
   \label{fig:qualitative_results}
\end{figure*}

\begin{table*}[t]
   \centering
   \small
   \setlength{\tabcolsep}{3.2pt}
   \renewcommand{\arraystretch}{1.08}
   \caption{\textsc{Stride} ablation study with ERNIE-Image as the base generator.
      We compare sampling-only baselines, execution routes, perception grounding, and refinement iterations on
      BizGenEval.}
   \label{tab:stride_ablation}
   \resizebox{\textwidth}{!}{%
      \begin{tabular}{l|ccccc|ccccccccc|c}
         \toprule
         $\#$      & \multicolumn{5}{c|}{Components} & \multicolumn{9}{c|}{BizGenEval} & \multicolumn{1}{c}
                                                                                         {\textbf{Avg.}}
         \\
         \cmidrule(lr){2-6} \cmidrule(lr){7-15} \cmidrule(lr){16-16}
                   & Lead                            & Comp.                           & Ret.
                   & Perc.
                   &
         Iters.    & Slides                          & Webpage                         & Poster
                   & Chart
                   &
         Sci. Fig. & Layout                          & Attr.                           & Text
                   & Know.
                   & \textbf{Score}
         \\
         \midrule

         (1)       & \multicolumn{4}{c}{Best-of-1}
                   &
         1 iter
                   & 32.8                            & 41.5                            & 35.6
                   & 10.5
                   &
         19.7
                   & 51.8                            & 31.8                            & 28.3
                   & 0.1
                   &
         28.0
         \\

         (2)       & \multicolumn{4}{c}{Best-of-6}
                   &
         1 iter
                   & 39.0                            & 48.5                            & 40.2
                   & 19.0
                   &
         26.8
                   & 60.6                            & 41.4                            & 36.4
                   & 0.4
                   &
         34.7
         \\
         \midrule

         (3)       & \checkmark                      & \checkmark                      & --
                   & --
                   &
         3 iters
                   & 48.5                            & 53.2                            & 44.8
                   & 21.2
                   &
         38.0
                   & 63.8                            & 45.4                            & 40.6
                   & 14.8
                   &
         41.1
         \\

         (4)       & \checkmark                      & --                              & \checkmark
                   & --
                   &
         3 iters
                   & 57.5                            & 60.5                            & 52.0
                   & 38.8
                   &
         42.5
                   & 64.4                            & 41.4                            & 60.4
                   & 34.8
                   &
         50.2
         \\

         (5)       & \checkmark                      & \checkmark                      & \checkmark
                   & --
                   &
         3 iters
                   & 57.0                            & 57.0                            & 54.5
                   & 34.8
                   &
         41.5
                   & 64.2                            & 49.4                            & 51.0
                   & 31.2
                   &
         48.9
         \\

         (6)       & \checkmark                      & --                              & \checkmark
                   & \checkmark
                   &
         3 iters
                   & 58.2                            & 61.8                            & 54.2
                   & 36.5
                   &
         41.5
                   & 63.6                            & 44.0                            & 59.0
                   & 35.2
                   &
         50.4
         \\
         \midrule

         (7)       & \checkmark                      & \checkmark                      & \checkmark
                   & \checkmark
                   &
         1 iter
                   & 54.2                            & 56.5                            & 51.2
                   & 30.8
                   &
         39.0
                   & 63.2                            & 45.8                            & 51.6
                   & 24.8
                   &
         46.4
         \\

         (8)       & \checkmark                      & \checkmark                      & \checkmark
                   & \checkmark
                   &
         2 iters
                   & 56.0                            & 59.0                            & 51.5
                   & 34.0
                   &
         40.2
                   & 63.6                            & 47.8                            & 53.8
                   & 27.4
                   &
         48.1
         \\
         \rowcolor{OursPink}
         (9)       & \checkmark                      & \checkmark                      & \checkmark
                   & \checkmark
                   &
         3 iters
                   & 61.3                            & 60.2                            & 54.5
                   & 37.0
                   &
         42.5
                   & 64.4                            & 50.8                            & 55.2
                   & 34.0
                   &
         51.1
         \\
         \bottomrule
      \end{tabular}%
   }
\end{table*}

\textbf{Qualitative Results.}
As shown in Figure~\ref{fig:qualitative_results}, compared with contemporary advanced agentic frameworks,
\textsc{Stride} produces infographics that better satisfy the specified constraints, exhibit more coherent layouts,
place semantic attributes more appropriately, and render text more accurately. These results meet the stringent
requirements of infographic generation and demonstrate clear advantages.
Additional qualitative results and a detailed case study are provided in Appendices~\ref{app:more_qualitative_results} and~\ref{app:case_study}.

\subsection{Ablation Studies}

Table~\ref{tab:stride_ablation} ablates the main components of \textsc{Stride} using ERNIE-Image as the base
generator. We report BizGenEval hard scores. Rows (1)--(2) compare against sampling-only baselines,
while the remaining rows progressively enable the lead editor, composer, retoucher, perceptor, and multiple refinement
iterations. Detailed results are provided in Appendix~\ref{app:extend_ablation_result}.

\textbf{Candidate Exploration.}
Increasing the candidate pool from Best-of-1 to Best-of-6 (Table~\ref{tab:stride_ablation} (1)-(2)) raises the BizGenEval average from 28.0 to 34.7, but the result remains far below the full system’s score of 51.1.
Notably, the knowledge score stays close to zero across all candidate-pool sizes.
Thus, the gains require structured diagnosis and targeted refinement, rather than mere best-of-$K$ sampling.

\textbf{Scale-Aware Execution.}
Composer-only (Table~\ref{tab:stride_ablation} (3):(2)) improves the average to 41.1, with a large gain on knowledge constraints over the base model.
Retoucher-only (Table~\ref{tab:stride_ablation} (4):(2)) is stronger, reaching 50.2 and substantially improving text score from 36.4 to 60.4.
The complete scale-aware system (Table~\ref{tab:stride_ablation} (9):(2)) achieves the best score of 51.1, which supports the scale-aware design.

\textbf{Dual-Route Evaluation.}
In particular, adding the perceptor to the composer–retoucher configuration (Table~\ref{tab:stride_ablation} (9):(5)) improves the three-iteration score from 48.9 to 51.1, suggesting that grounding is useful when it supports route selection and local editing jointly.

\textbf{Iterative Refinement.}
With all components enabled, performance increases from 46.4 to 48.1 and 51.1 across one, two, and three iterations (Table~\ref{tab:stride_ablation} (7)-(8)-(9)), showing that the refinement loop accumulates improvements rather than merely adding stochastic variation.

\textbf{Evolution Strategy.}
Table~\ref{tab:evolution_strategy_main} shows that both broader candidate exploration and incumbent retention contribute to our $(1+\lambda)$-evolution strategy. Increasing the candidate set ($\lambda$) from one to four raises the average score from 39.4 to 45.4 under forced replacement, and from 45.6 to 51.1 when retaining the incumbent, indicating that a larger candidate pool improves the chance of finding a strong proposal. Moreover, at the same candidate budget, incumbent retention yields gains of 6.2 points for $(1+1)$ over $(1,1)$ and 5.7 points for $(1+4)$ over $(1,4)$, as it avoids discarding the current champion for a verifier-non-improving candidate.
More detailed analysis in Appendix~\ref{app:evolution_strategy}.

\begin{table*}[t]
   \centering
   \scriptsize
   \setlength{\tabcolsep}{2.4pt}
   \renewcommand{\arraystretch}{1.08}

   \begin{minipage}[t]{0.36\textwidth}
      \centering
      \captionof{table}{Comparison of champion-evolution strategies on BizGenEval.
         The $(1,1)$ and $(1,4)$ variants denote forced replacement, whereas
         the $(1+1)$ and $(1+4)$ variants retain the incumbent champion unless
         a better candidate is found.}
      \label{tab:evolution_strategy_main}
      \resizebox{\linewidth}{!}{%
         \begin{tabular}{@{}l|cccc|c@{}}
            \toprule
            \multirow{2}{*}{\textbf{Strategy}}
                                                       & \multicolumn{4}{c|}{\textbf{Capability}}
                                                       & \multirow{2}{*}{\textbf{Average}}                                               \\
            \cmidrule(lr){2-5}
                                                       & Layout                                   & Attr. & Text & Know. &               \\
            \midrule
            Random selection                           & 54.6                                     & 40.0  & 46.4 & 20.6  & 40.4          \\
            $(1,1)$ forced replacement                 & 56.0                                     & 37.8  & 42.2 & 21.4  & 39.4          \\
            $(1+1)$-evolution strategy                 & 62.6                                     & 43.6  & 49.2 & 26.8  & 45.6          \\
            $(1,4)$ forced replacement                 & 59.2                                     & 42.4  & 50.4 & 29.6  & 45.4          \\
            \rowcolor{OursPink}
            \textbf{$(1+4)$-evolution strategy (ours)} & 64.4                                     & 50.8  & 55.2 & 34.0  & \textbf{51.1} \\
            \bottomrule
         \end{tabular}%
      }
   \end{minipage}
   \hfill
   \begin{minipage}[t]{0.62\textwidth}
      \centering
      \captionof{table}{Human and automated evaluation of the paired subset from
         BizGenEval and IGenBench. The
         frozen generator and \textsc{Stride} outputs are evaluated using human
         consensus (H), the Qwen3.5-397B-A17B verifier (Q), and the corresponding
         official benchmark evaluator (O).}
      \label{tab:human_mllm_combined}
      \resizebox{\linewidth}{!}{%
         \begin{tabular}{@{}ll|ccc|ccc|ccc|cc@{}}
            \toprule
            \multirow{2}{*}{\textbf{Benchmark}}
                       & \multirow{2}{*}{\textbf{Generator}}
                       & \multicolumn{3}{c|}{\textbf{Frozen generator}}
                       & \multicolumn{3}{c|}{\textbf{\textsc{Stride}}}
                       & \multicolumn{3}{c|}{\textbf{$\Delta$ (\textsc{Stride} $-$ Frozen)}}
                       & \multicolumn{2}{c}{\textbf{Correlation}}                                                                                                                              \\
            \cmidrule(lr){3-5}
            \cmidrule(lr){6-8}
            \cmidrule(lr){9-11}
            \cmidrule(l){12-13}
                       &
                       & \textbf{H}                                                          & \textbf{Q}            & \textbf{O}
                       & \textbf{H}                                                          & \textbf{Q}            & \textbf{O}
                       & \textbf{H}                                                          & \textbf{Q}            & \textbf{O}
                       & $\rho_{\mathrm{H-Q}}$                                               & $\rho_{\mathrm{H-O}}$                                                                           \\
            \midrule
            \multicolumn{13}{c}{\textbf{BizGenEval: Hard Score}}                                                                                                                               \\
            \midrule
            BizGenEval & ERNIE-Image
                       & 30.1                                                                & 39.0                  & 32.8       & 50.4 & 63.8 & 53.8 & +20.3 & +24.8 & +21.0 & 0.650 & 0.732 \\
            BizGenEval & Infographic-V2
                       & 41.5                                                                & 40.0                  & 42.4       & 71.3 & 77.6 & 70.6 & +29.8 & +37.6 & +28.2 & 0.725 & 0.811 \\
            \midrule
            \multicolumn{13}{c}{\textbf{IGenBench: Q-ACC}}                                                                                                                                     \\
            \midrule
            IGenBench  & ERNIE-Image
                       & 0.42                                                                & 0.50                  & 0.43       & 0.58 & 0.66 & 0.57 & +0.16 & +0.16 & +0.14 & 0.708 & 0.733 \\
            IGenBench  & Infographic-V2
                       & 0.76                                                                & 0.82                  & 0.77       & 0.88 & 0.93 & 0.89 & +0.12 & +0.11 & +0.12 & 0.785 & 0.847 \\
            \bottomrule
         \end{tabular}%
      }
   \end{minipage}
\end{table*}

\subsection{Human-MLLM Studies.}
\label{sec:human_mllm}
We examine whether automated evaluation aligns with human judgments
and whether the gains of \textsc{Stride} persist under human evaluation (Appendix~\ref{app:human_mllm_setup}).
Table~\ref{tab:human_mllm_combined} shows positive
correlations with human scores for both the Qwen verifier
($\rho=0.650$--$0.785$) and the official evaluators
($\rho=0.732$--$0.847$).
Human evaluation also preserves the improvement of \textsc{Stride}
in all four configurations: BizGenEval hard score increases by
20.3 and 29.8 points for ERNIE-Image and Infographic-V2, respectively,
while IGenBench Q-ACC increases by 0.16 and 0.12.
These results support human alignment of the automated rankings and
provide independent evidence for the refinement gains.

\section{Conclusion}

We presented \textsc{Stride}, a training-free multi-agent framework that formulates infographic generation as scale-aware evolutionary refinement over frozen text-to-image generators.
\textsc{Stride} separates rubric-based verification from spatial grounding, routes failures to either global composition or local retouching according to their scale, and admits only candidates that improve verifier-scored constraint satisfaction under an evolution strategy.
Across BizGenEval, IGenBench and WISE, \textsc{Stride} consistently improves open-weight generators, outperforms prior agentic enhancement methods, and matches or surpasses proprietary systems.
We hope this work inspires future advancements in agentic infographic generation and refinement.

\subsection*{AI use statement}

Large Language Models (LLMs) were used in this study to aid in polishing the manuscript. Specifically, we used LLMs to assist in refining the language and detecting potential grammatical errors. This is to improve readability and ensure clarity of the paper. We confirm that LLMs were not involved in research ideation, method exploration, and experiment designs. All research ideas, methods, and analysis were produced by the authors. We take full responsibility for the content in this paper, including the text generated or polished by the LLMs.

\bibliography{iclr2027_conference}
\bibliographystyle{iclr2027_conference}

\appendix
\clearpage

\startcontents[appendix]
\section*{Appendix Contents}
\printcontents[appendix]{}{1}{}

\clearpage

\section{Rubric Checklist Instantiation}
\label{app:constraint_interface}

\textsc{Stride} requires a set of atomic rubrics that can be evaluated on an image. This appendix describes how
that set is obtained on each benchmark. On BizGenEval and IGenBench we take it from the released checklist. On WISE, which provides no checklist, \textsc{Stride} constructs it dynamically from the prompt with the prompt parser $\Phi$.

\textbf{BizGenEval.}
Each prompt is paired with 20 verification questions, 10 easy and 10 hard, constructed from the prompt through a
human-in-the-loop process. The released annotation procedure requires every question to be unambiguous, visually
verifiable, and consistent with the prompt specification.
We use the 20 questions as $\mathcal{C}$ and
instantiate the two priority groups of the hard and easy subsets.

\textbf{IGenBench.}
The question set for a prompt $p$ is $\mathcal{Q}(p)=\mathcal{Q}_p(p)\cup\mathcal{Q}_e(p)$. The prompt-derived part
$\mathcal{Q}_p$ is obtained by splitting the prompt into sentences and converting each sentence into a self-contained
yes/no question, and accounts for 3{,}730 of the 5{,}259 released questions. The expert-informed part $\mathcal{Q}_e$
instantiates three general visualization requirements, namely data completeness, data ordering, and data encoding,
and accounts for the remaining 1{,}529. We use $\mathcal{Q}(p)$ as $\mathcal{C}$ with a single priority group.

\textbf{WISE.}
WISE provides no checklist, so we instantiate the prompt parser as $\mathcal{C}=\Phi(P)$, where $\Phi$ prompts Kimi-K3 to decompose the prompt into atomic, image-answerable requirements in a structured format.
The prompt parser is instantiated with the following prompt.

\begin{promptbox}[Prompt Parser Prompt]{CigarBlue}{CigarBlueBg}
   KIMI_ENHANCE_SYSTEM = """You are a world knowledge expert specializing in visual content understanding.

   Your task: Given a short prompt that may contain implicit knowledge (like cultural references, historical events, scientific concepts), infer the intended subject and provide an enhanced prompt suitable for image generation.

   IMPORTANT: You must rely on your own knowledge. Do NOT reveal that you're using search engines.
   """

   KIMI_ENHANCE_USER_TEMPLATE = """Original prompt: "{original_prompt}"

   Please provide (output as JSON):
   {{
         "reasoning": "Your detailed thought process - what knowledge is required to understand this prompt",
         "confidence": 0.0-1.0,  // Your confidence in the reasoning
         "enhanced_prompt": "Explicit, visual description for image generation (can be a complete rewrite)",
         "verification_checklist": {{
               "primary_subject": "Main object/concept that MUST appear",
               "key_features": ["feature1", "feature2", ...],
               "optional_context": "Optional scene/atmosphere"
            }}
      }}

   Rules:
   1. If confidence < 0.7, include "need_search" in your reasoning
   2. Enhanced prompt should be explicit and visual (no implicit knowledge)
   3. Verification checklist is YOUR inference (not ground truth - you don't have it)
   4. You can completely rewrite the prompt if needed
   """
\end{promptbox}

\section{Benchmark Details}
\label{app:benchmarks}

\begin{table*}[!htbp]
   \centering
   \footnotesize
   \setlength{\tabcolsep}{4pt}
   \renewcommand{\arraystretch}{1.14}
   \caption{Quantitative overview of the public benchmarks used in our evaluation. Final reporting follows each benchmark's official evaluator and metrics.}
   \label{tab:benchmark_overview}
   \begin{tabularx}{\textwidth}{l c X c l}
      \toprule
      \textbf{Benchmark} & \textbf{Scale} & \textbf{Coverage}                             & \textbf{Final eval. units} & \textbf{Metrics} \\
      \midrule
      BizGenEval         & 400 prompts    & 5 domains $\times$ 4 capabilities = 20 tasks  & 8,000 checklist judgments  & Score, Accuracy  \\
      IGenBench          & 600 cases      & 30 infographic types in 6 categories          & 5,259 QA judgments         & Q-ACC, I-ACC     \\
      WISE               & 1,000 prompts  & 3 domains, 25 subdomains, 6 report categories & 3,000 dimension scores     & WiScore          \\
      \bottomrule
   \end{tabularx}
\end{table*}

We evaluate \textsc{Stride} on three large-scale public benchmarks that provide complementary tests of infographic and structured visual generation. BizGenEval focuses on commercial visual documents with dense text, layout, attribute, and knowledge constraints. IGenBench targets text-to-infographic reliability, especially faithful data encoding and end-to-end constraint satisfaction. WISE evaluates knowledge-grounded generation from indirect semantic cues. Table~\ref{tab:benchmark_overview} summarizes the scale, coverage, and evaluation protocol of each benchmark.

\textbf{BizGenEval.}
BizGenEval~\citep{li2026bizgeneval} crosses five professional content domains (slides, charts, webpages, posters, and scientific figures) with four capability dimensions of text rendering, layout control, attribute binding, and knowledge-based reasoning.
The prompt set contains 300 content-based prompts derived from real-world commercial visual materials and 100 knowledge-based prompts covering physics, chemistry, mathematics, history, and arts. Each prompt is paired with 20 task-specific verification questions, and final scores are computed from MLLM checklist judgments on easy and hard tracks.
BizGenEval uses Gemini-3-Flash-Preview~\citep{googledeepmindGemini3Flash} as its official evaluator.

BizGenEval's official code reports two closely related checklist-based metrics. Let $q$ denote the number of questions in a track and
let $Q_{\mathrm{err}}$ be the number of incorrectly answered questions. The (1) \textit{checklist accuracy} is
\begin{equation}
   \mathrm{Accuracy}=\frac{q-Q_{\mathrm{err}}}{q}.
\end{equation}
The official leaderboard (2) \textit{error score} applies a penalty to the number of errors,
\begin{equation}
   \mathrm{Score}=\max\bigl(0, 1-\alpha Q_{\mathrm{err}}\bigr),
\end{equation}
where BizGenEval sets $\alpha=0.2$. Thus, for a 10-question track, the score becomes zero once five or more questions
are answered incorrectly.
Following the leaderboard protocol, BizGenEval-related results report the official error score, unless otherwise stated.

\textbf{IGenBench.}
IGenBench~\citep{tang2026igenbench} contains 600 curated test cases spanning 30 infographic types,
its evaluation decomposes each infographic specification into atomic yes/no verification questions.
These questions cover 10 types of infographic elements (Title \& Subtitle, Chart/Diagram Type, Decorative/Non-data Elements, Annotations \& Callouts, Axes \& Scales, Legend \& Category Mapping, Data Marks, Data Completeness, Data Ordering, and Data Encoding).
IGenBench uses Gemini-2.5-Pro~\citep{comanici2025gemini} as its official evaluator, reporting both Q-ACC, which measures individual constraint satisfaction, and I-ACC, which requires all questions for an infographic to be answered correctly.
The official released dataset contains 5,259 atomic yes/no verification questions, including 3,730 prompt-derived questions and 1,529 expert-informed seed questions.

\textbf{WISE.}
WISE~\citep{niu2025wise} evaluates world-knowledge-informed image generation. Unlike benchmarks that use direct prompts, WISE avoids naming the target entity explicitly and instead describes it through clues, relations, or conditions.
WISE contains 1,000 prompts across 25 subdomains in three major areas, including 400 cultural prompts, 300 spatio-temporal prompts, and 300 natural-science prompts, while the latter two are further divided into time, space, biology, physics, and chemistry categories.
WISE uses WiScore as its metric and uses GPT-4o-2024-05-13~\citep{hurst2024gpt} as its official evaluator. WiScore combines consistency, realism, and aesthetic quality, with default weights of 0.7, 0.2, and 0.1, respectively, and normalizes the weighted score to $[0,1]$.

\section{Advantages over Existing Agentic Methods}
\label{app:agentic_comparison}

\begin{table*}[!htbp]
   \centering
   \scriptsize
   \setlength{\tabcolsep}{2.6pt}
   \renewcommand{\arraystretch}{1.06}
   \caption{Comparison of agentic methods on the mechanisms required for infographic refinement in the inference time. A, G and V for
      the agent backbone, base generator, and verifier, with O, P, and -- denoting open-source or open-weight,
      proprietary, and not applicable.}
   \label{tab:agent_comparison}
   \resizebox{\textwidth}{!}{%
      \begin{tabular}{@{}l c c c c c c c c@{}}
         \toprule
         \textbf{Method}          & \textbf{Venue}           & \textbf{A / G / V}         & \textbf{WebSearch}
                                  & \textbf{Rubric Feedback} & \textbf{Global Redraw}
                                  & \textbf{Local Edit}      & \textbf{Iterative Refine.} & \textbf{Champion Admit}                                                                        \\
         \midrule
         ReflectionFlow           & 25@ICCV                  & -- / O / O                 & $\times$                & $\times$         & $\surd$          & $\times$ & $\surd$  & $\times$ \\
         T2I-Copilot              & 25@ICCV                  & P / O / O                  & $\times$                & $\times$         & $\surd$          & $\times$ & $\surd$  & $\times$ \\
         Ag3                      & 25@ICML                  & P / P / P                  & $\times$                & $\times$         & $\surd$          & $\times$ & $\surd$  & $\times$ \\
         Gen-Searcher             & 26@Arxiv                 & O / P / --                 & $\surd$                 & $\times$         & $\times$         & $\times$ & $\times$ & $\times$ \\
         GenEvolve                & 26@Arxiv                 & O / P / --                 & $\surd$                 & $\times$         & $\times$         & $\times$ & $\times$ & $\times$ \\
         Mind-Brush               & 26@SIGAsia               & P / P / --                 & $\surd$                 & $\times$         & $\times$         & $\times$ & $\times$ & $\times$ \\
         PromptEnhancer           & 26@CVPR                  & O / O / O                  & $\times$                & $\times$         & $\times$         & $\times$ & $\times$ & $\times$ \\
         PaperBanana              & 26@ICML                  & P / P / P                  & $\surd$                 & $\times$         & $\surd$          & $\times$ & $\surd$  & $\times$ \\
         AutoFigure               & 26@ICLR                  & P / P / P                  & $\times$                & $\times$         & $\surd$          & $\surd$  & $\surd$  & $\times$ \\
         \rowcolor{OursPink}
         \textbf{\textsc{Stride}} & \textbf{Ours}            & \textbf{O / O / O}         & \textbf{$\surd$}
                                  & \textbf{$\surd$}         & \textbf{$\surd$}           & \textbf{$\surd$}        & \textbf{$\surd$} & \textbf{$\surd$}                                  \\
         \bottomrule
      \end{tabular}%
   }
\end{table*}

Existing agentic methods can be partitioned into the following three categories:

(1) \textbf{Retrieval-Augmented Planning Methods.} A line of retrieval-augmented planning methods, exemplified by PromptEnhancer~\citep{wang2025promptenhancer}, Mind-Brush~\citep{he2026mind}, Gen-Searcher~\citep{feng2026gen}, and GenEvolve~\citep{chen2026genevolve}, augments text-to-image generation by searching for implicit prompt content or retrieving reference images to condition the generator. These methods subsequently reorganize the prompt through a planning stage, yielding an enhanced prompt intended to improve generation quality.
Nevertheless, these approaches are fundamentally limited to one-shot prompt rewriting. As a result, they not only fall short of enforcing the desired constraints, but can also compromise previously verified correct content.

(2) \textbf{Layout-Augmented Planning Methods.} Methods such as Ag3~\citep{hahn2024proactive} and AutoFigure~\citep{zhu2026autofigure} augment generation with explicit layout planning, typically via editable graphs or SVGs, followed by fixed-round layout-guided refinement.
However, their reliance on closed-source frontier models throughout generation and optimization obscures whether the gains stem from the agentic design or from the underlying models. Their extra layout-control components also limit compatibility with most open-source text-to-image generators.

(3) \textbf{Analysis-Evaluation Methods.} Methods represented by ReflectionFlow~\citep{zhuo2025reflection}, T2I-Copilot~\citep{chen2025t2i}, and PaperBanana~\citep{zhu2026paperbanana} refine generated images under the guidance of a verifier that evaluates the current output. Yet the verifier feedback is typically not rubric-grounded, making it difficult to translate detected failures into precise and actionable corrections. In addition, their execution space is narrow, usually restricted to prompt revision or full regeneration.
As a result, full-image regeneration can introduce regressions by altering already verified correct regions, leaving limited room for reliable improvement.

These existing systems improve over single-pass generation in many settings, but dense infographic generation requires additional mechanisms.
In particular, a refinement loop must identify which rubrics failed, decide whether the failure should be addressed by global regeneration or localized editing, and prevent later revisions from degrading constraints that were already satisfied.

\textsc{Stride} addresses these limitations and demonstrates clear advantages. We compare these agentic methods in Table~\ref{tab:agent_comparison}.
For each representative method, we record whether it uses web search, rubric feedback as refinement state, global redrawing, local editing, iterative refinement, and champion admission, and we report the access profile of the agent backbone, base generator, and verifier.

For \textsc{Stride}, all components are open-weight, reducing attribution ambiguity and making the reported gains reflect the framework rather than proprietary model strength.
The failed rubrics returned by dual-route evaluation guide retrieval, planning, and editing, allowing each refinement step to address the concrete errors present in the current image.
During iterations, \textsc{Stride} operates in a scale-aware execution space, adaptively combining global regeneration with local editing so that the two serve complementary roles rather than conflicting alternatives.
Finally, a $(1+\lambda)$-evolution strategy accepts a proposal only when it strictly improves the current champion, ensuring monotonic and non-regressive refinement across rounds.

\begin{table*}[!htbp]
   \centering
   \scriptsize
   \setlength{\tabcolsep}{2.1pt}
   \renewcommand{\arraystretch}{0.95}
   \caption{Extended quantitative comparison on BizGenEval.
      Each entry reports the hard score followed by the easy score in parentheses.
      Rows are ordered by Venue and, within each year, by hard Average.
      Params denotes the parameters of the base generator.
      The proprietary-model results are reproduced unchanged from the official
      table of BizGenEval. Missing entries are denoted by ``--''.}
   \label{tab:bizgenevalextend}

   \resizebox{\textwidth}{!}{%
      \begin{tabular}{@{}l | c | c | *{9}{c} | c @{}}
         \toprule
         \multirow{2}{*}{\textbf{Model}}
                                                   & \multirow{2}{*}{\textbf{Venue}}
                                                   & \multirow{2}{*}{\textbf{Params}}
                                                   & \multicolumn{5}{c}{\textbf{Visual Domains}}
                                                   & \multicolumn{4}{c|}{\textbf{Capabilities}}
                                                   & \multirow{2}{*}{\textbf{Average}}
         \\
         \cmidrule(lr){4-8}
         \cmidrule(lr){9-12}
                                                   &                                             &
                                                   & \textbf{Slides}                             & \textbf{Webpage}
                                                   & \textbf{Poster}
                                                   & \textbf{Chart}                              &
         \shortstack{\textbf{Sci.}\\\textbf{Figure}}
                                                   & \textbf{Layout}                             & \textbf{Attribute}
                                                   & \textbf{Text}
                                                   & \textbf{Knowledge}                          &
         \\
         \midrule
         \rowcolor{GroupGray}
         \multicolumn{13}{l}{\textit{Proprietary models}}
         \\

         Nano-Banana-Pro                           & 25@Google                                   & --
                                                   & 82.2 (94.8)
                                                   & 77.5 (96.5)
                                                   & 76.5 (94.8)
                                                   & 73.0 (92.2)
                                                   & 74.2 (90.0)
                                                   & 72.2 (91.2)
                                                   & 65.6 (92.2)
                                                   & 86.4 (95.0)
                                                   & 82.6 (96.2)
                                                   & 76.7 (93.7)
         \\

         Wan2.6-T2I                                & 26@Alibaba                                  & --
                                                   & 27.1 (56.5)                                 & 25.5 (67.0)
                                                   & 27.5 (62.5)
                                                   & 17.2 (48.8)                                 & 12.5 (58.8)
                                                   & 46.4 (80.6)                                 & 16.6 (60.6)
                                                   & 12.6 (52.6)
                                                   & 12.2 (41.0)                                 & 21.9 (58.7)
         \\

         Seedream-5.0                              & 26@ByteDance                                & --
                                                   & 54.5 (80.8)                                 & 47.0 (80.8)
                                                   & 50.7 (77.0)
                                                   & 46.0 (76.2)                                 & 45.8 (81.5)
                                                   & 67.6 (89.0)                                 & 42.4 (77.2)
                                                   & 43.4 (75.6)
                                                   & 41.8 (75.2)                                 & 48.8 (79.2)
         \\

         Seedream-5.0-Pro                          & 26@ByteDance                                & --
                                                   & --                                          & --
                                                   & --
                                                   & --                                          & --
                                                   & 75.5 (87.6)                                 & 74.1 (87.3)
                                                   & 63.2 (81.1)
                                                   & 49.8 (75.3)                                 & 65.7 (82.8)
         \\

         GPT-Image-1.5                             & 26@OpenAI                                   & --
                                                   & 40.8 (89.2)                                 & 41.0 (86.0)
                                                   & 42.0 (83.5)
                                                   & 28.2 (76.5)                                 & 27.8 (72.8)
                                                   & 51.6 (84.8)                                 & 25.8 (75.2)
                                                   & 40.4 (82.8)
                                                   & 26.0 (83.6)                                 & 35.9 (81.6)
         \\

         Nano-Banana-2.0                           & 26@Google                                   & --
                                                   & 73.8 (95.8)                                 & 71.2 (94.5)
                                                   & 67.5 (91.2)
                                                   & 60.2 (89.2)                                 & 69.5
         (92.0)
                                                   & 68.4 (91.0)                                 & 57.4 (91.6)
                                                   & 83.4 (94.6)
                                                   & 64.6 (93.0)                                 & 68.5 (92.5)
         \\

         \midrule
         \rowcolor{GroupGray}
         \multicolumn{13}{l}{\textit{Open-weight models}}
         \\

         Qwen-Image                                & 25@Alibaba                                  & 20B
                                                   & 3.5 (28.5)                                  & 2.5 (27.5)
                                                   & 5.8 (32.2)
                                                   & 2.2 (15.8)                                  & 0.0 (15.2)
                                                   & 10.4 (51.2)                                 & 0.2 (22.2)
                                                   & 0.6 (17.6)
                                                   & 0.0 (4.4)                                   & 2.8 (23.8)
         \\

         FLUX.2-dev                                & 25@BFL                                      & 32B
                                                   & 5.5 (43.2)                                  & 5.5 (48.8)
                                                   & 7.2 (48.2)
                                                   & 5.5 (33.5)                                  & 0.5 (36.5)
                                                   & 17.2 (67.8)                                 & 1.2 (49.2)
                                                   & 1.0 (43.0)
                                                   & 0.0 (8.2)                                   & 4.9 (42.0)
         \\

         Qwen-Image-2512                           & 25@Alibaba                                  & 20B
                                                   & 10.2 (45.0)                                 & 7.2 (48.0)
                                                   & 11.5 (47.0)
                                                   & 2.2 (28.0)                                  & 0.2 (37.0)
                                                   & 22.2 (70.6)                                 & 1.2 (47.8)
                                                   & 1.8 (39.2)
                                                   & 0.0 (6.4)                                   & 6.3 (41.0)
         \\

         Z-Image                                   & 25@Z-Image                                  & 6B
                                                   & 12.2 (43.8)                                 & 6.2 (48.5)
                                                   & 12.0 (50.0)
                                                   & 8.8 (30.5)                                  & 1.8 (46.0)
                                                   & 26.8 (69.2)                                 & 2.6 (47.6)
                                                   & 2.8 (45.0)
                                                   & 0.6 (13.2)                                  & 8.2 (43.8)
         \\

         HunyuanImage-3.0                          & 25@Tencent                                  & 80B-A13B
                                                   & 19.3 (47.0)                                 & 21.0 (52.0)
                                                   & 19.8 (53.5)
                                                   & 2.0 (18.8)                                  & 2.8 (29.0)
                                                   & 27.8 (65.0)                                 & 13.8 (53.6)
                                                   & 10.2 (39.6)
                                                   & 0.0 (2.0)                                   & 13.0 (40.1)
         \\

         Emu3.5                                    & 25@BAAI                                     & 34.1B
                                                   & 14.5 (44.8)                                 & 20.0 (48.8)
                                                   & 20.3 (53.0)
                                                   & 4.8 (20.3)                                  & 6.5 (34.0)
                                                   & 30.4 (63.4)                                 & 14.2 (52.6)
                                                   & 7.0 (33.6)
                                                   & 1.2 (11.0)                                  & 13.2 (40.2)
         \\

         ERNIE-Image                               & 26@Baidu                                    & 8B
                                                   & 32.8 (57.0)                                 & 41.5 (65.6)
                                                   & 35.6 (62.8)
                                                   & 10.5 (35.3)                                 & 19.7 (55.4)
                                                   & 51.8 (78.7)                                 & 31.8 (73.9)
                                                   & 28.3 (65.1)
                                                   & 0.1 (3.1)                                   & 28.0 (55.2)
         \\

         Infographic-V2                            & 26@SenseTime                                & 8B
                                                   & 41.5 (64.2)                                 & 45.5 (66.8)
                                                   & 43.5 (63.7)
                                                   & 41.5 (59.2)                                 & 33.5 (64.8)
                                                   & 58.8 (91.4)                                 & 42.6 (73.8)
                                                   & 63.0 (84.4)
                                                   & 0.0 (5.4)                                   & 41.1 (63.7)
         \\

         \midrule
         \rowcolor{GroupGray}
         \multicolumn{13}{l}{\textit{Agentic framework}}
         \\

         PE (ERNIE-Image)                          & 26@Baidu                                    & 8B-3B
                                                   & 30.5 (58.2)                                 & 33.5 (59.5)
                                                   & 30.0 (50.7)
                                                   & 11.8 (40.0)                                 & 15.2 (46.2)
                                                   & 39.2 (56.6)                                 & 27.4 (61.8)
                                                   & 26.0 (60.4)
                                                   & 4.2 (25.0)                                  & 24.2 (50.9)
         \\

         GenEvolve (ERNIE-Image)                   & 26@Arxiv                                    & 8B
                                                   & 26.8 (48.5)                                 & 39.3 (63.8)
                                                   & 28.3 (49.0)
                                                   & 9.8 (39.0)                                  & 10.3 (41.3)
                                                   & 42.8 (65.6)                                 & 27.0 (58.2)
                                                   & 18.8 (54.6)
                                                   & 2.8 (14.8)                                  & 22.9 (48.3)
         \\

         GenEvolve (ERNIE-Image-PE)                & 26@Arxiv                                    & 8B-3B
                                                   & 25.5 (52.5)                                 & 30.8 (52.8)
                                                   & 24.2 (50.0)
                                                   & 9.8 (36.8)                                  & 11.8 (46.8)
                                                   & 35.6 (59.6)                                 & 20.4 (59.2)
                                                   & 19.4 (50.6)
                                                   & 6.2 (21.6)                                  & 20.4 (47.8)
         \\

         Gen-Searcher (ERNIE-Image)                & 26@Arxiv                                    & 8B
                                                   & 31.8 (63.5)                                 & 36.5 (63.0)
                                                   & 37.2 (65.0)
                                                   & 11.0 (40.8)                                 & 22.0 (60.5)
                                                   & 47.6 (78.0)                                 & 30.2 (67.6)
                                                   & 23.6 (62.6)
                                                   & 9.4 (26.0)                                  & 27.7 (58.6)
         \\

         Gen-Searcher (Infographic-V2)             & 26@Arxiv                                    & 8B
                                                   & 46.5 (76.2)                                 & 41.0 (71.8)
                                                   & 45.0 (67.8)
                                                   & 41.8 (62.7)                                 & 41.8 (74.2)
                                                   & 58.2 (85.4)                                 & 38.6 (75.6)
                                                   & 58.2 (78.8)
                                                   & 17.8 (42.4)                                 & 43.2 (70.5)
         \\

         PromptEnhancer-7B (ERNIE-Image)           & 26@CVPR                                     & 8B
                                                   & 15.5 (42.8)                                 & 17.2 (46.8)
                                                   & 14.8 (42.2)
                                                   & 9.2 (21.2)                                  & 3.2 (23.2)
                                                   & 32.4 (65.4)                                 & 9.0 (44.0)
                                                   & 6.2 (26.0)
                                                   & 0.4 (5.6)                                   & 12.0 (35.2)
         \\

         PromptEnhancer-32B (ERNIE-Image)          & 26@CVPR                                     & 8B
                                                   & 18.5 (47.8)                                 & 17.5 (49.8)
                                                   & 18.0 (51.7)
                                                   & 12.0 (37.8)                                 & 5.5 (39.0)
                                                   & 35.4 (73.4)                                 & 12.6 (55.8)
                                                   & 8.0 (41.8)
                                                   & 1.2 (9.8)                                   & 14.3 (45.2)
         \\

         PromptEnhancer-7B (Infographic-V2)        & 26@CVPR                                     & 8B
                                                   & 13.5 (41.8)                                 & 9.5 (34.0)
                                                   & 13.2 (37.5)
                                                   & 10.0 (24.2)                                 & 2.8 (27.3)
                                                   & 30.6 (64.8)                                 & 5.8 (38.2)
                                                   & 2.2 (18.6)
                                                   & 0.6 (10.2)                                  & 9.8 (33.0)
         \\

         PromptEnhancer-32B (Infographic-V2)       & 26@CVPR                                     & 8B
                                                   & 17.2 (46.5)                                 & 14.2 (42.8)
                                                   & 12.8 (49.0)
                                                   & 19.0 (44.2)                                 & 6.8 (41.2)
                                                   & 36.0 (80.6)                                 & 10.0 (49.0)
                                                   & 8.6 (38.0)
                                                   & 1.4 (11.4)                                  & 14.0 (44.8)
         \\

         \rowcolor{OursPink}
         \textbf{\textsc{Stride} (ERNIE-Image)}    & \textbf{Ours}                               & 8B
                                                   & \underline{61.3 (79.2)}                     & \underline{60.2 (83.2)}
                                                   & \underline{54.5 (74.5)}
                                                   & \underline{37.0 (62.3)}                     & \underline{42.5 (69.2)}
                                                   & \underline{64.4 (80.8)}                     & \underline{50.8 (82.6)}
                                                   & \underline{55.2 (79.8)}
                                                   & \underline{34.0 (51.6)}                     & \underline{51.1 (73.7)}
         \\
         \rowcolor{OursPink}
         \textbf{\textsc{Stride} (Infographic-V2)} & \textbf{Ours}                               & 8B
                                                   & \textbf{73.2 (86.2)}                        & \textbf{69.5 (88.8)}
                                                   & \textbf{74.5 (85.5)}
                                                   & \textbf{67.5 (78.2)}                        & \textbf{63.0 (82.2)}
                                                   & \textbf{79.2 (96.0)}                        & \textbf{65.8 (85.6)}
                                                   & \textbf{79.2 (88.4)}
                                                   & \textbf{54.0 (66.8)}                        & \textbf{69.5 (84.2)}
         \\

         \bottomrule
      \end{tabular}%
   }
\end{table*}

\section{Extended Experiments}

\subsection{Extended Main Results}
\label{app:extend_main_results}

Table~\ref{tab:bizgenevalextend} extends the BizGenEval results by reporting both hard and easy scores for each visual domain, capability, and their average. This provides a more comprehensive view of \textsc{Stride}'s performance: the hard track reflects performance on more demanding requirements, while the easy track complements it by showing how well each method handles less challenging cases.

On the hard track, \textsc{Stride} consistently improves its base generator across all reported domains and capabilities. With ERNIE-Image, \textsc{Stride} raises the average score from 28.0 to 51.1, a gain of 23.1 points. With Infographic-V2, it raises the score from 41.1 to 69.5, a gain of 28.4 points. The improvements also extend to the easy track, where the corresponding averages increase from 55.2 to 73.7 (+18.5) and from 63.7 to 84.2 (+20.5), respectively. For example, with Infographic-V2, the hard scores for layout, attribute, text, and knowledge improve by 20.4, 23.2, 16.2, and 54.0 points, respectively. These results show that the gains are not confined to visual presentation, but also include capabilities related to satisfying specified attributes, rendering text, and incorporating relevant knowledge.

\textsc{Stride} also substantially outperforms the listed prior agentic frameworks when paired with the same base generator. On the hard track, its average exceeds the strongest corresponding baseline by 23.4 points with ERNIE-Image (51.1 vs.\ 27.7) and by 26.3 points with Infographic-V2 (69.5 vs.\ 43.2). The advantage is also present on the easy track, with gains of 15.1 points (73.7 vs.\ 58.6) and 13.7 points (84.2 vs.\ 70.5), respectively. Thus, the improvements are not limited to a single difficulty track or a single base generator.

Finally, \textsc{Stride} (Infographic-V2) surpasses several proprietary models on the hard average, including Seedream-5.0 (48.8), Seedream-5.0-Pro (65.7), and Nano-Banana-2.0 (68.5), with a score of 69.5. It also exceeds these models on the easy average where applicable: its score of 84.2 is higher than Seedream-5.0 (79.2) and Seedream-5.0-Pro (82.8), though lower than Nano-Banana-2.0 (92.5). These comparisons indicate that an open-weight generator combined with \textsc{Stride} can outperform a number of proprietary systems, particularly on the more demanding hard track.

\begin{table*}[!htbp]
   \centering
   \small
   \setlength{\tabcolsep}{3.2pt}
   \renewcommand{\arraystretch}{1.08}
   \caption{Extended \textsc{Stride} ablation study on BizGenEval and WISE benchmarks.}
   \label{tab:stride_ablation_extended}
   \resizebox{\textwidth}{!}{%
      \begin{tabular}{l|ccccc|ccccccccc|c|c}
         \toprule
         $\#$        & \multicolumn{5}{c|}{Components} & \multicolumn{9}{c|}{BizGenEval} & \multicolumn{1}{c|}
                                                                                           {\textbf{Average}} & \multicolumn{1}{c}{WISE}
         \\
         \cmidrule(lr){2-6} \cmidrule(lr){7-15} \cmidrule(lr){16-16} \cmidrule(lr){17-17}
                     & Lead                            & Comp.                           & Ret.                &
         Perc.
                     &
         Iters.      & Slides                          & Webpage                         & Poster              &
         Chart
                     &
         Sci. Fig.   & Layout                          & Attr.                           & Text                &
         Know.
                     & \textbf{Score}                  & Overall
         \\
         \midrule

         (1)         & \multicolumn{4}{c}{Best-of-1}
                     &
         1 iter
                     & 32.8 (57.0)                     & 41.5 (65.6)                     & 35.6 (62.8)         &
         10.5 (35.3)
                     &
         19.7 (55.4)
                     & 51.8 (78.7)                     & 31.8 (73.9)                     & 28.3 (65.1)         &
         0.1 (3.1)
                     &
         28.0 (55.2) & 0.58
         \\

         (2)         & \multicolumn{4}{c}{Best-of-6}
                     &
         1 iter
                     & 39.0 (58.2)                     & 48.5 (69.0)                     & 40.2 (64.8)         &
         19.0 (44.8)
                     &
         26.8 (58.0)
                     & 60.6 (81.0)                     & 41.4 (80.4)                     & 36.4 (70.6)         &
         0.4 (3.8)
                     &
         34.7 (59.0) & --
         \\
         \midrule

         (3)         & \checkmark                      & \checkmark                      & --
                     & --
                     &
         1 iter
                     & 44.5 (65.0)                     & 53.5 (71.2)                     & 45.5 (64.5)         &
         21.2 (46.5)
                     &
         37.5 (69.0)
                     & 63.8 (81.6)                     & 43.0 (80.4)                     & 39.4 (69.2)         &
         15.6 (21.8)
                     &
         40.4 (63.2) & 0.89
         \\

         (4)         & \checkmark                      & \checkmark                      & --
                     & --
                     &
         2 iters
                     & 44.5 (64.0)                     & 54.2 (70.5)                     & 44.5 (66.0)         &
         22.2 (48.8)
                     &
         38.7 (69.8)
                     & 64.4 (81.6)                     & 43.2 (80.6)                     & 40.8 (68.8)         &
         15.0 (24.2)
                     &
         40.8 (63.8) & --
         \\
         \rowcolor{GroupGray}
         (5)         & \checkmark                      & \checkmark                      & --
                     & --
                     &
         3 iters
                     & 48.5 (66.5)                     & 53.2 (69.5)                     & 44.8 (65.0)         &
         21.2 (49.0)
                     &
         38.0 (68.8)
                     & 63.8 (80.6)                     & 45.4 (78.4)                     & 40.6 (70.0)         &
         14.8 (26.0)
                     &
         41.1 (63.7) & --
         \\
         \midrule

         (6)         & \checkmark                      & --                              & \checkmark
                     & --
                     &
         1 iter
                     & 52.5 (74.2)                     & 59.2 (80.8)                     & 51.0 (74.2)         &
         34.0 (59.0)
                     &
         37.8 (67.8)
                     & 63.4 (81.2)                     & 40.8 (80.8)                     & 53.8 (78.8)         &
         29.6 (44.0)
                     &
         46.9 (71.2) & --
         \\

         (7)         & \checkmark                      & --                              & \checkmark
                     & --
                     &
         2 iters
                     & 55.2 (76.8)                     & 59.0 (82.2)                     & 52.0 (76.0)         &
         36.5 (61.3)
                     &
         39.8 (69.5)
                     & 62.2 (82.2)                     & 40.4 (81.4)                     & 57.4 (80.2)         &
         34.0 (48.8)
                     &
         48.5 (73.2) & --
         \\
         \rowcolor{GroupGray}
         (8)         & \checkmark                      & --                              & \checkmark
                     & --
                     &
         3 iters
                     & 57.5 (79.8)                     & 60.5 (82.2)                     & 52.0 (76.5)         &
         38.8 (63.0)
                     &
         42.5 (74.0)
                     & 64.4 (83.6)                     & 41.4 (80.8)                     & 60.4 (83.6)         &
         34.8 (52.4)
                     &
         50.2 (75.1) & --
         \\
         \midrule

         (9)         & \checkmark                      & \checkmark                      & \checkmark
                     & --
                     &
         1 iter
                     & 52.8 (73.2)                     & 55.5 (77.0)                     & 51.5 (71.0)         &
         30.8 (52.8)
                     &
         37.8 (66.2)
                     & 63.2 (81.8)                     & 45.6 (81.2)                     & 49.4 (75.6)         &
         24.4 (33.6)
                     &
         45.6 (68.0) & 0.89
         \\

         (10)        & \checkmark                      & \checkmark                      & \checkmark
                     & --
                     &
         2 iters
                     & 55.5 (73.2)                     & 57.2 (81.2)                     & 51.7 (72.8)         &
         31.0 (55.0)
                     &
         39.0 (68.2)
                     & 63.4 (81.8)                     & 46.6 (80.8)                     & 50.4 (75.8)         &
         27.2 (42.0)
                     &
         46.9 (70.1) & --
         \\
         \rowcolor{GroupGray}
         (11)        & \checkmark                      & \checkmark                      & \checkmark
                     & --
                     &
         3 iters
                     & 57.0 (74.2)                     & 57.0 (82.2)                     & 54.5 (74.0)         &
         34.8 (59.8)
                     &
         41.5 (68.0)
                     & 64.2 (80.8)                     & 49.4 (82.4)                     & 51.0 (77.2)         &
         31.2 (46.2)
                     &
         48.9 (71.7) & --
         \\
         \midrule

         (12)        & \checkmark                      & --                              & \checkmark          &
         \checkmark
                     &
         1 iter
                     & 52.8 (74.0)                     & 59.2 (81.0)                     & 52.0 (74.0)         &
         33.5 (58.0)
                     &
         37.8 (68.5)
                     & 64.2 (81.0)                     & 40.8 (80.2)                     & 53.4 (78.6)         &
         29.8 (44.6)
                     &
         47.0 (71.1) & --
         \\

         (13)        & \checkmark                      & --                              & \checkmark          &
         \checkmark
                     &
         2 iters
                     & 54.8 (76.5)                     & 59.8 (83.0)                     & 54.2 (75.8)         &
         35.0 (61.3)
                     &
         38.8 (68.5)
                     & 61.6 (81.8)                     & 43.0 (81.2)                     & 55.4 (80.4)         &
         34.0 (48.6)
                     &
         48.5 (73.0) & --
         \\
         \rowcolor{GroupGray}
         (14)        & \checkmark                      & --                              & \checkmark          &
         \checkmark
                     &
         3 iters
                     & 58.2 (80.0)                     & 61.8 (83.8)                     & 54.2 (76.2)         &
         36.5 (62.5)
                     &
         41.5 (73.0)
                     & 63.6 (83.2)                     & 44.0 (80.4)                     & 59.0 (84.0)         &
         35.2 (52.8)
                     &
         50.4 (75.1) & --
         \\
         \midrule

         (15)        & \checkmark                      & \checkmark                      & \checkmark          &
         \checkmark
                     &
         1 iter
                     & 54.2 (74.0)                     & 56.5 (77.2)                     & 51.2 (71.0)         &
         30.8 (52.8)
                     &
         39.0 (67.2)
                     & 63.2 (81.8)                     & 45.8 (81.2)                     & 51.6 (76.4)         &
         24.8 (34.4)
                     &
         46.4 (68.5) & 0.91
         \\

         (16)        & \checkmark                      & \checkmark                      & \checkmark          &
         \checkmark
                     &
         2 iters
                     & 56.0 (74.5)                     & 59.0 (82.0)                     & 51.5 (72.8)         &
         34.0 (57.0)
                     &
         40.2 (69.2)
                     & 63.6 (81.8)                     & 47.8 (81.0)                     & 53.8 (77.6)         &
         27.4 (44.0)
                     &
         48.1 (71.1) & 0.92
         \\
         \rowcolor{OursPink}
         (17)        & \checkmark                      & \checkmark                      & \checkmark          &
         \checkmark
                     &
         3 iters
                     & 61.3 (79.2)                     & 60.2 (83.2)                     & 54.5 (74.5)         &
         37.0 (62.3)
                     &
         42.5 (69.2)
                     & 64.4 (80.8)                     & 50.8 (82.6)                     & 55.2 (79.8)         &
         34.0 (51.6)
                     &
         51.1 (73.7) & 0.92
         \\
         \bottomrule
      \end{tabular}%
   }
\end{table*}

\begin{table*}[t]
   \centering
   \small
   \setlength{\tabcolsep}{9pt}
   \renewcommand{\arraystretch}{1.2}
   \caption{Per-iteration scale-aware sub-agent preference of the lead editor, reported as
      composer / retoucher percentages on the BizGenEval with two generators. The number of rounds is
      given in parentheses.}
   \label{tab:analyzer_iter}
   \resizebox{0.9\textwidth}{!}{%
      \begin{tabular}{@{}ll cccc@{}}
         \toprule
         \textbf{Setup} & \textbf{Dimension} & \textbf{Iter 1}     & \textbf{Iter 2}     & \textbf{Iter 3}     & \textbf{Overall}             \\
         \midrule
         \multirow{5}{*}{ERNIE-Image}
                        & Layout             & 97.2 / 2.8 ($72$)   & 92.7 / 7.3 ($55$)   & 89.6 / 10.4 ($48$)  & 93.7 / 6.3 ($175$)           \\
                        & Attribute          & 82.7 / 17.3 ($81$)  & 77.3 / 22.7 ($75$)  & 81.9 / 18.1 ($72$)  & 80.7 / 19.3 ($228$)          \\
                        & Text               & 9.2 / 90.8 ($87$)   & 17.6 / 82.4 ($68$)  & 23.7 / 76.3 ($59$)  & 15.9 / 84.1 ($214$)          \\
                        & Knowledge          & 55.1 / 44.9 ($98$)  & 53.4 / 46.6 ($88$)  & 42.2 / 57.8 ($83$)  & 50.6 / 49.4 ($269$)          \\
         \cmidrule(l){2-6}
                        & \textit{Total}     & 58.9 / 41.1 ($338$) & 58.7 / 41.3 ($286$) & 57.6 / 42.4 ($262$) & \textbf{58.5 / 41.5 ($886$)} \\
         \midrule
         \multirow{5}{*}{Infographic-V2}
                        & Layout             & 94.4 / 5.6 ($72$)   & 93.4 / 6.6 ($61$)   & 96.2 / 3.8 ($52$)   & 94.6 / 5.4 ($185$)           \\
                        & Attribute          & 82.7 / 17.3 ($81$)  & 82.7 / 17.3 ($75$)  & 85.9 / 14.1 ($71$)  & 83.7 / 16.3 ($227$)          \\
                        & Text               & 11.5 / 88.5 ($87$)  & 15.5 / 84.5 ($71$)  & 31.6 / 68.4 ($57$)  & 18.1 / 81.9 ($215$)          \\
                        & Knowledge          & 71.6 / 28.4 ($95$)  & 68.3 / 31.7 ($82$)  & 74.2 / 25.8 ($66$)  & 71.2 / 28.8 ($243$)          \\
         \cmidrule(l){2-6}
                        & \textit{Total}     & 63.6 / 36.4 ($335$) & 64.4 / 35.6 ($289$) & 72.4 / 27.6 ($246$) & \textbf{66.3 / 33.7 ($870$)} \\
         \bottomrule
      \end{tabular}
   }
\end{table*}

\subsection{Extended Ablation Results}
\label{app:extend_ablation_result}

Table~\ref{tab:stride_ablation_extended} provides a more detailed view of the component and iteration ablations, reporting both hard and easy scores across BizGenEval dimensions, together with WISE scores (WiScore) where available. These results complement the main-text ablation by showing how performance evolves under different component configurations and refinement budgets.

\begin{table*}[!htbp]
   \centering
   \scriptsize
   \setlength{\tabcolsep}{2.1pt}
   \renewcommand{\arraystretch}{0.95}
   \caption{Extended quantitative comparison on BizGenEval with oracle experiments.
      Each entry reports the hard score followed by the easy score in parentheses.
      Rows are ordered by Venue and, within each year, by hard average.
      Params denotes the parameters of the base generator.
      The proprietary-model results are reproduced unchanged from the official
      table of BizGenEval benchmark. Missing entries are denoted by ``--''.}
   \label{tab:bizgenevaloracle}

   \resizebox{\textwidth}{!}{%
      \begin{tabular}{@{}l | c | c | *{9}{c} | c @{}}
         \toprule
         \multirow{2}{*}{\textbf{Model}}
                                                   & \multirow{2}{*}{\textbf{Venue}}
                                                   & \multirow{2}{*}{\textbf{Params}}
                                                   & \multicolumn{5}{c}{\textbf{Visual Domains}}
                                                   & \multicolumn{4}{c|}{\textbf{Capabilities}}
                                                   & \multirow{2}{*}{\textbf{Average}}
         \\
         \cmidrule(lr){4-8}
         \cmidrule(lr){9-12}
                                                   &                                             &
                                                   & \textbf{Slides}                             & \textbf{Webpage}
                                                   & \textbf{Poster}
                                                   & \textbf{Chart}                              &
         \shortstack{\textbf{Sci.}\\\textbf{Figure}}
                                                   & \textbf{Layout}                             & \textbf{Attribute}
                                                   & \textbf{Text}
                                                   & \textbf{Knowledge}                          &
         \\
         \midrule
         \rowcolor{GroupGray}
         \multicolumn{13}{l}{\textit{Proprietary models}}
         \\

         Nano-Banana-Pro                           & 25@Google                                   & --
                                                   & 82.2 (94.8)
                                                   & \underline{77.5 (96.5)}
                                                   & \underline{76.5 (94.8)}
                                                   & \underline{73.0 (92.2)}
                                                   & \textbf{74.2 (90.0)}
                                                   & 72.2 (91.2)
                                                   & 65.6 (92.2)
                                                   & \underline{86.4 (95.0)}
                                                   & \textbf{82.6 (96.2)}
                                                   & \underline{76.7 (93.7)}
         \\

         Wan2.6-T2I                                & 26@Alibaba                                  & --
                                                   & 27.1 (56.5)                                 & 25.5 (67.0)
                                                   & 27.5 (62.5)
                                                   & 17.2 (48.8)                                 & 12.5 (58.8)
                                                   & 46.4 (80.6)                                 & 16.6 (60.6)
                                                   & 12.6 (52.6)
                                                   & 12.2 (41.0)                                 & 21.9 (58.7)
         \\

         Seedream-5.0                              & 26@ByteDance                                & --
                                                   & 54.5 (80.8)                                 & 47.0 (80.8)
                                                   & 50.7 (77.0)
                                                   & 46.0 (76.2)                                 & 45.8 (81.5)
                                                   & 67.6 (89.0)                                 & 42.4 (77.2)
                                                   & 43.4 (75.6)
                                                   & 41.8 (75.2)                                 & 48.8 (79.2)
         \\

         Seedream-5.0-Pro                          & 26@ByteDance                                & --
                                                   & --                                          & --
                                                   & --
                                                   & --                                          & --
                                                   & 75.5 (87.6)                                 & \textbf{74.1 (87.3)}
                                                   & 63.2 (81.1)
                                                   & 49.8 (75.3)                                 & 65.7 (82.8)
         \\

         GPT-Image-1.5                             & 26@OpenAI                                   & --
                                                   & 40.8 (89.2)                                 & 41.0 (86.0)
                                                   & 42.0 (83.5)
                                                   & 28.2 (76.5)                                 & 27.8 (72.8)
                                                   & 51.6 (84.8)                                 & 25.8 (75.2)
                                                   & 40.4 (82.8)
                                                   & 26.0 (83.6)                                 & 35.9 (81.6)
         \\

         Nano-Banana-2.0                           & 26@Google                                   & --
                                                   & \underline{73.8 (95.8)}                     & 71.2 (94.5)
                                                   & 67.5 (91.2)
                                                   & 60.2 (89.2)                                 & 69.5
         (92.0)
                                                   & 68.4 (91.0)                                 & 57.4 (91.6)
                                                   & 83.4 (94.6)
                                                   & 64.6 (93.0)                                 & 68.5 (92.5)
         \\

         \midrule
         \rowcolor{GroupGray}
         \multicolumn{13}{l}{\textit{Open-weight models}}
         \\

         ERNIE-Image                               & 26@Baidu                                    & 8B
                                                   & 32.8 (57.0)                                 & 41.5 (65.6)
                                                   & 35.6 (62.8)
                                                   & 10.5 (35.3)                                 & 19.7 (55.4)
                                                   & 51.8 (78.7)                                 & 31.8 (73.9)
                                                   & 28.3 (65.1)
                                                   & 0.1 (3.1)                                   & 28.0 (55.2)
         \\

         Infographic-V2                            & 26@SenseTime                                & 8B
                                                   & 41.5 (64.2)                                 & 45.5 (66.8)
                                                   & 43.5 (63.7)
                                                   & 41.5 (59.2)                                 & 33.5 (64.8)
                                                   & 58.8 (91.4)                                 & 42.6 (73.8)
                                                   & 63.0 (84.4)
                                                   & 0.0 (5.4)                                   & 41.1 (63.7)
         \\

         \midrule
         \rowcolor{GroupGray}
         \multicolumn{13}{l}{\textit{Agentic framework}}
         \\

         \rowcolor{OursPink}
         \textbf{\textsc{Stride} (ERNIE-Image)}    & \textbf{Ours}                               & 8B
                                                   & 61.3 (79.2)                                 & 60.2 (83.2)
                                                   & 54.5 (74.5)
                                                   & 37.0 (62.3)                                 & 42.5 (69.2)
                                                   & 64.4 (80.8)                                 & 50.8 (82.6)
                                                   & 55.2 (79.8)
                                                   & 34.0 (51.6)                                 & 51.1 (73.7)
         \\
         \rowcolor{OursPink}
         \textbf{\textsc{Stride} (Infographic-V2)} & \textbf{Ours}                               & 8B
                                                   & 73.2 (86.2)                                 & 69.5 (88.8)
                                                   & 74.5 (85.5)
                                                   & 67.5 (78.2)                                 & 63.0 (82.2)
                                                   & \underline{79.2 (96.0)}                     & 65.8 (85.6)
                                                   & 79.2 (88.4)
                                                   & 54.0 (66.8)                                 & 69.5 (84.2)
         \\

         \midrule
         \rowcolor{GroupGray}
         \multicolumn{13}{l}{\textit{Oracle}}
         \\

         \textbf{\textsc{Stride} (ERNIE-Image)}    & \textbf{Ours}                               & 8B
                                                   & 70.2 (87.8)                                 & 76.0 (93.2)
                                                   & 70.5 (86.8)
                                                   & 47.8 (68.0)                                 & 57.5 (82.5)
                                                   & 71.8 (87.4)
                                                   & 63.6 (86.2)
                                                   & 67.2 (88.2)
                                                   & 54.8 (72.8)                                 & 64.4 (83.7)
         \\

         \textbf{\textsc{Stride} (Infographic-V2)} & \textbf{Ours}                               & 8B
                                                   & \textbf{81.2 (91.5)}                        & \textbf{80.8 (95.2)}
                                                   & \textbf{78.8 (89.2)}
                                                   & \textbf{75.0 (87.2)}                        & \underline{70.5 (90.2)}
                                                   & \textbf{85.2 (97.2)}
                                                   & \underline{73.2 (90.4)}
                                                   & \textbf{88.4 (94.2)}
                                                   & \underline{62.2 (81.0)}                     & \textbf{77.2 (90.7)}
         \\

         \bottomrule
      \end{tabular}%
   }
\end{table*}

\begin{table*}[t]
   \centering
   \scriptsize
   \setlength{\tabcolsep}{2.4pt}
   \renewcommand{\arraystretch}{0.95}
   \caption{Extended quantitative comparison on IGenBench with oracle experiments.
      Params denotes the parameters of the base generator.
      The proprietary-model results are reproduced unchanged from the official IGenBench.}
   \label{tab:igenbench_results_oracle}

   \resizebox{\textwidth}{!}{%
      \begin{tabular}{@{}l | c | c | c c c c c c c c c c | c c@{}}
         \toprule
         \multirow{2}{*}{\textbf{Model}}
                          & \multirow{2}{*}{\textbf{Venue}}
                          & \multirow{2}{*}{\textbf{Params}}
                          & \multicolumn{10}{c|}{\textbf{Question Type Q-ACC}}
                          & \multicolumn{2}{c}{\textbf{Overall}}
         \\
         \cmidrule(lr){4-13}
         \cmidrule(lr){14-15}
                          &                                                    &                  &
         \textbf{Comp.}   & \textbf{Enc.}                                      & \textbf{Order}
                          & \textbf{Marks}                                     & \textbf{Anno.}   &
         \textbf{Axes}
                          & \textbf{Leg.}                                      & \textbf{Chart}   &
         \textbf{Title}   & \textbf{Deco.}
                          & \textbf{Q-ACC}                                     & \textbf{I-ACC}
         \\
         \midrule

         \rowcolor{GroupGray}
         \multicolumn{15}{l}{\textit{Proprietary models}}
         \\

         Image-01         & 25@MiniMax                                         & --
                          & 0.01                                               & 0.05             & 0.04
                          & 0.10                                               & 0.10
                          & 0.14                                               & 0.03             & 0.22
                          & 0.14                                               & 0.47             & 0.13          & 0.00 \\

         P-Image          & 25@PrunaAI                                         & --
                          & 0.08                                               & 0.15             & 0.19
                          & 0.27                                               & 0.36
                          & 0.28                                               & 0.54             & 0.43
                          & 0.58                                               & 0.68             & 0.34          & 0.00 \\

         Nano-Banana      & 25@Google                                          & --
                          & 0.18                                               & 0.31             & 0.27
                          & 0.44                                               & 0.54
                          & 0.57                                               & 0.52             & 0.60
                          & 0.65                                               & 0.81             & 0.48          & 0.02 \\

         Nano-Banana-Pro  & 25@Google                                          & --
                          & \underline{0.84}                                   & \textbf{0.86}    &
         \underline{0.90}
                          & \textbf{0.87}                                      & \underline{0.93} &
         \underline{0.93}
                          & \textbf{0.96}                                      & \underline{0.92} &
         \underline{0.98}
                          & \underline{0.94}                                   & \textbf{0.90}    &
         \textbf{0.49}                                                                                                   \\

         GPT-Image-1.5    & 26@OpenAI                                          & --
                          & 0.38                                               & 0.48             & 0.44
                          & 0.57                                               & 0.50             & 0.54
                          & 0.57                                               & 0.68             & 0.60
                          & 0.80                                               & 0.55             & 0.12                 \\

         Seedream-4.5     & 26@ByteDance                                       & --
                          & 0.34                                               & 0.37             & 0.47
                          & 0.48                                               & 0.70             & 0.70
                          & 0.81                                               & 0.68             & 0.95
                          & 0.84                                               & 0.61             & 0.06                 \\

         Seedream-5.0-Pro & 26@ByteDance                                       & --
                          & 0.64                                               & 0.47             & 0.71
                          & 0.73                                               & 0.77             & 0.85
                          & 0.83                                               & 0.83             & 0.86
                          & 0.88                                               & 0.73             & 0.17                 \\

         GPT-Image-2.0    & 26@OpenAI                                          & --
                          & \textbf{0.89}                                      & 0.61             & \textbf{0.93}
                          & \textbf{0.87}                                      & \textbf{0.94}    & \textbf{0.98}
                          & \underline{0.95}                                   & \textbf{0.93}    & \textbf{0.99}
                          & \textbf{0.95}                                      & \underline{0.88} & 0.38                 \\

         Qwen-Image-3.0   & 26@Alibaba                                         & --
                          & 0.65                                               & 0.43             & 0.77
                          & 0.77                                               & 0.84             & 0.84
                          & 0.87                                               & 0.85             & 0.80
                          & 0.88                                               & 0.74             & 0.13                 \\
         \midrule
         \rowcolor{GroupGray}
         \multicolumn{15}{l}{\textit{Open-weight models}}
         \\

         ERNIE-Image      & 26@Baidu                                           & 8B
                          & 0.17                                               & 0.16             & 0.25
                          & 0.34                                               & 0.43
                          & 0.47                                               & 0.61             & 0.55
                          & 0.57                                               & 0.81             & 0.41          & 0.01 \\

         Infographic-V2   & 26@SenseTime                                       & 8B
                          & 0.54                                               & 0.49             & 0.69
                          & 0.66                                               & 0.79
                          & 0.80                                               & 0.85             & 0.80
                          & 0.94                                               & 0.88
                          & 0.72                                               & 0.17
         \\

         \midrule
         \rowcolor{GroupGray}
         \multicolumn{15}{l}{\textit{Agentic framework}}
         \\

         \rowcolor{OursPink}
         \textbf{\textsc{Stride} (ERNIE-Image)}
                          & \textbf{Ours}                                      & 8B
                          & 0.30                                               & 0.32             & 0.41
                          & 0.52                                               & 0.60
                          & 0.63                                               & 0.73             & 0.70
                          & 0.67                                               & 0.89
                          & 0.55                                               & 0.03
         \\

         \rowcolor{OursPink}
         \textbf{\textsc{Stride} (Infographic-V2)}
                          & \textbf{Ours}                                      & 8B
                          & 0.70                                               & 0.72             & 0.81
                          & 0.78                                               & 0.90
                          & 0.89                                               & 0.92             & 0.91
                          & \underline{0.98}                                   & \underline{0.94}
                          & 0.84                                               & 0.38
         \\

         \midrule
         \rowcolor{GroupGray}
         \multicolumn{15}{l}{\textit{Oracle}}
         \\

         \textbf{\textsc{Stride} (Infographic-V2)}
                          & \textbf{Ours}                                      & 8B
                          & 0.71                                               & \underline{0.74} &
         0.82
                          & \underline{0.80}                                   & 0.92
                          & 0.90                                               & 0.92             &
         \underline{0.92}
                          & \textbf{0.99}                                      & \underline{0.94}
                          & 0.85                                               & \underline{0.40}
         \\

         \bottomrule
      \end{tabular}%
   }
\end{table*}

\textbf{Benefits of Combining Components and Iterating.}
Combining the composer and retoucher yields higher hard average scores than using the composer alone at each matched iteration budget: 45.6 vs.\ 40.4 after one iteration (Table~\ref{tab:stride_ablation_extended} (9):(3)), 46.9 vs.\ 40.8 after two (Table~\ref{tab:stride_ablation_extended} (10):(4)), and 48.9 vs.\ 41.1 after three (Table~\ref{tab:stride_ablation_extended} (11):(5)). Adding the perceptor to the lead-editor--retoucher configuration also improves the three-iteration hard average score from 50.2 to 50.4 (Table~\ref{tab:stride_ablation_extended} (14):(8)), with gains in slides, webpage, and scientific figure scores. With all components enabled (Table~\ref{tab:stride_ablation_extended} (15)-(16)-(17)), the hard average score increases from 46.4 after one iteration to 48.1 after two and 51.1 after three, and the corresponding Easy Average increases from 68.5 to 71.1 and 73.7.
The WISE rows show transfer beyond checklist benchmarks, \textsc{Stride} improves the score from 0.58 under the Best-of-1 baseline to 0.92 after two iterations (Table~\ref{tab:stride_ablation_extended} (16):(1)), further demonstrating the efficiency of the \textsc{Stride} framework.

\textbf{Scale-Aware Sub-Agent Preferences.}
Table~\ref{tab:analyzer_iter} further breaks down the lead editor's preferences for the composer and retoucher by generator, dimension, and iteration. The preferences vary substantially across dimensions, rather than consistently favoring a single sub-agent. For both generators, the composer is strongly preferred for layout (93.7\% overall for ERNIE-Image and 94.6\% for Infographic-V2) and is also preferred for attribute and knowledge. In contrast, the retoucher is preferred for text, accounting for 84.1\% of the decisions with ERNIE-Image and 81.9\% with Infographic-V2. This pattern is consistent with the different strengths of the two routes: broader composition changes are more often selected for layout-level issues, while localized editing is more often selected for text-related corrections.

The preference distributions also vary across iterations and generators. For ERNIE-Image, the overall composer preference remains relatively stable across the three rounds (58.9\%, 58.7\%, and 57.6\%), while for Infographic-V2 it rises from 63.6\% in the first round to 72.4\% in the third. The dimension-level trends are not uniform: for example, the retoucher remains the majority choice for text in every round for both generators, whereas the composer remains strongly preferred for layout. These differences indicate that the lead editor's routing behavior depends on the generator, the error dimension, and the refinement round, rather than following a fixed composer-only or retoucher-only policy. Together with the improved results of the combined system, these preference patterns provide evidence for the scale-aware operation of \textsc{Stride}.

\subsection{Extended Oracle Experiment Results}
\label{app:oracle_experiment_results}

\begin{table*}[!htbp]
   \centering
   \scriptsize
   \setlength{\tabcolsep}{2.4pt}
   \renewcommand{\arraystretch}{0.95}
   \caption{Quantitative comparison on WISE. We report WiScore across six knowledge domains and the overall score.
      Best and second-best results are shown in bold and underline, respectively. The proprietary-model results are reproduced unchanged from the official WISE.}
   \label{tab:wise_results}

   \resizebox{\textwidth}{!}{%
      \begin{tabular}{@{}l | c | c | c c c c c c | c@{}}
         \toprule
         \textbf{Model}                             & \textbf{Venue}   & \textbf{Params}
                                                    & \textbf{Cult.}   & \textbf{Time}    & \textbf{Space}
                                                    & \textbf{Bio.}    & \textbf{Phys.}   & \textbf{Chem.}   & \textbf{Overall}                      \\
         \midrule
         \rowcolor{GroupGray}
         \multicolumn{10}{l}{\textit{Proprietary models}}                                                                                            \\

         Seedream 4.0                               & 25@ByteDance     & --
                                                    & 0.78             & 0.73             & 0.85             & 0.79             & 0.84 & 0.67 & 0.78 \\

         GPT-4o / GPT-Image-1                       & 25@OpenAI        & --
                                                    & 0.81             & 0.71             & \underline{0.89} & 0.83             & 0.79 & 0.74 & 0.80 \\

         Nano-Banana-Pro                            & 25@Google        & --
                                                    & 0.89             & 0.80             & \underline{0.89}
                                                    & 0.88             & 0.86             & 0.85             & 0.87                                  \\

         \midrule
         \rowcolor{GroupGray}
         \multicolumn{10}{l}{\textit{Open-weight models}}                                                                                            \\

         FLUX.1-dev                                 & 24@BFL           & 12B
                                                    & 0.48             & 0.58             & 0.62             & 0.42             & 0.51 & 0.35 & 0.50 \\

         Hunyuan-Image 3.0                          & 25@Tencent       & 80B-A13B
                                                    & 0.58             & 0.57             & 0.70             & 0.56             & 0.63 & 0.31 & 0.57 \\

         Qwen-Image                                 & 25@Alibaba       & 20B
                                                    & 0.62             & 0.63             & 0.77             & 0.57             & 0.75 & 0.40 & 0.62 \\

         LongCat-Image                              & 25@Meituan       & 6B
                                                    & 0.66             & 0.61             & 0.72             & 0.66             & 0.72 & 0.49 & 0.65 \\

         BAGEL+CoT                                  & 25@ByteDance     & 7B
                                                    & 0.76             & 0.69             & 0.75             & 0.65             & 0.75 & 0.58 & 0.70 \\

         NextFlow-RL                                & 26@NextFlow      & --
                                                    & 0.63             & 0.63             & 0.77             & 0.58             & 0.67 & 0.39 & 0.62 \\

         SenseNova-U1-SFT                           & 26@SenseTime     & 8B
                                                    & 0.65             & 0.66             & 0.82             & 0.68             & 0.81 & 0.66 & 0.69 \\

         NEO-unify-CoT                              & 26@SenseTime     & 8B
                                                    & 0.73             & 0.67             & 0.79             & 0.70             & 0.75 & 0.66 & 0.72 \\

         DeepGen1.0                                 & 26@DeepGen       & --
                                                    & 0.72             & 0.81             & 0.70             & 0.67             & 0.82 & 0.66 & 0.73 \\

         ERNIE-Image                                & 26@Baidu         & 8B
                                                    & 0.54             & 0.59             & 0.78             & 0.55             & 0.66 & 0.35 & 0.58 \\

         SenseNova-U1-SFT                           & 26@SenseTime     & 8B-A3B
                                                    & 0.73             & 0.69             & 0.80             & 0.73             & 0.82 & 0.69 & 0.74 \\

         SenseNova-U1-SFT-CoT                       & 26@SenseTime     & 8B
                                                    & 0.78             & 0.73             & 0.82             & 0.80             & 0.85 & 0.85 & 0.78 \\

         SenseNova-U1-SFT-CoT                       & 26@SenseTime     & 8B-A3B
                                                    & 0.81             & 0.77             & 0.84             & 0.81             & 0.84 & 0.82 & 0.81 \\

         \midrule
         \rowcolor{GroupGray}
         \multicolumn{10}{l}{\textit{Agentic framework}}                                                                                             \\

         GenAgent (FLUX.1-dev)                      & 26@arXiv         & 12B
                                                    & 0.78             & 0.67             & 0.78             & 0.72             & 0.77 & 0.55 & 0.72 \\

         Gen-Searcher (Qwen-Image)                  & 26@arXiv         & 20B
                                                    & 0.80             & 0.71             & 0.82             & 0.76             & 0.74 & 0.75 & 0.77 \\

         Mind-Brush (Qwen-Image-Edit-2511)          & 26@SIGGRAPH Asia & 20B
                                                    & 0.83             & 0.69             & 0.84             & 0.71             & 0.85 & 0.68 & 0.78 \\

         GenEvolve (Qwen-Image-Edit)                & 26@arXiv         & --
                                                    & 0.84             & 0.74             & 0.87             & 0.83             & 0.81 & 0.83 & 0.82 \\

         PE (ERNIE-Image)                           & 26@Baidu         & --
                                                    & 0.61             & 0.65             & 0.81             & 0.59             & 0.74 & 0.59 & 0.65 \\

         PromptEnhancer-32B (ERNIE-Image)           & 26@CVPR          & 8B
                                                    & 0.81             & 0.77             & 0.87             & 0.86             & 0.85 & 0.74 & 0.81 \\
         \rowcolor{OursPink}
         \textbf{\textsc{Stride} (Qwen-Image-2512)} & Ours             & 20B
                                                    & \textbf{0.93}    & \underline{0.84} & \textbf{0.93}
                                                    & \underline{0.89} & \underline{0.94} & \underline{0.90} & \underline{0.91}                      \\
         \rowcolor{OursPink}
         \textbf{\textsc{Stride} (ERNIE-Image)}     & Ours             & 8B
                                                    & \underline{0.92} & \textbf{0.87}    & \textbf{0.93}
                                                    & \textbf{0.92}    & \textbf{0.95}    & \textbf{0.92}    & \textbf{0.92}                         \\

         \bottomrule
      \end{tabular}%
   }
\end{table*}

To assess the model adaptability and further improvement potential of \textsc{Stride}, we replace only its perceptor and verifier with the proprietary models used by each benchmark, while keeping the open-weight generators and other framework components unchanged.
Specifically, in oracle experiments, \textsc{Stride} uses Gemini-3-Flash-Preview~\citep{googledeepmindGemini3Flash} as the perceptor and verifier on BizGenEval and Gemini-2.5-Pro~\citep{comanici2025gemini} for these roles on IGenBench, while all other components remain open-weight.

As shown in Tables~\ref{tab:bizgenevaloracle} and~\ref{tab:igenbench_results_oracle}, this stronger model configuration further improves both generators on BizGenEval: the hard average score increases from 51.1 to 64.4 with ERNIE-Image and from 69.5 to 77.2 with Infographic-V2. On IGenBench, using Infographic-V2, it also raises Q-ACC/I-ACC from 0.84/0.38 to 0.85/0.40. These results show that \textsc{Stride}'s refinement process can benefit from different perception and verification backbones across benchmarks and generators, revealing additional headroom for the same open-weight generators without changing the framework itself. We view this headroom as evidence of \textsc{Stride}'s extensibility: improvements in the models guiding refinement can translate into further gains from existing open-weight generators.

\subsection{Extended Model Scalability Results on WISE}
\label{app:wise_benchmark_result}

We further evaluate \textsc{Stride} on WISE to test whether the proposed refinement framework scales beyond infographic benchmarks with released checklists. Unlike BizGenEval and IGenBench, WISE does not provide explicit verification rubrics. Therefore, \textsc{Stride} constructs image-answerable constraints from the prompt using the parser $\Phi$ (Appendix~\ref{app:constraint_interface}) and applies the same verifier-guided refinement procedure.

\begin{table}[t]
   \centering
   \small
   \setlength{\tabcolsep}{4pt}
   \renewcommand{\arraystretch}{1.08}
   \caption{Sampling protocol for the human--MLLM evaluation study. For
      each prompt, we retain paired outputs from the frozen generator and
      \textsc{Stride} under the same generator and prompt.}
   \label{tab:human_mllm_protocol}
   \begin{tabular}{@{}l l c c l@{}}
      \toprule
      \textbf{Benchmark}
       & \textbf{Generator}
       & \textbf{Prompt pairs}
       & \textbf{Images}
       & \textbf{Human rubric}         \\
      \midrule
      BizGenEval
       & ERNIE-Image
       & 25
       & 50
       & $25 \times 10$ Hard questions \\

      BizGenEval
       & Infographic-V2
       & 25
       & 50
       & $25 \times 10$ Hard questions \\

      IGenBench
       & ERNIE-Image
       & 25
       & 50
       & All released atomic questions \\

      IGenBench
       & Infographic-V2
       & 25
       & 50
       & All released atomic questions \\
      \midrule

      \textbf{Total}
       & --
       & \textbf{100}
       & \textbf{200}
       & --                            \\
      \bottomrule
   \end{tabular}
\end{table}

\begin{figure*}[!htbp]
   \centering
   \includegraphics[width=\textwidth]{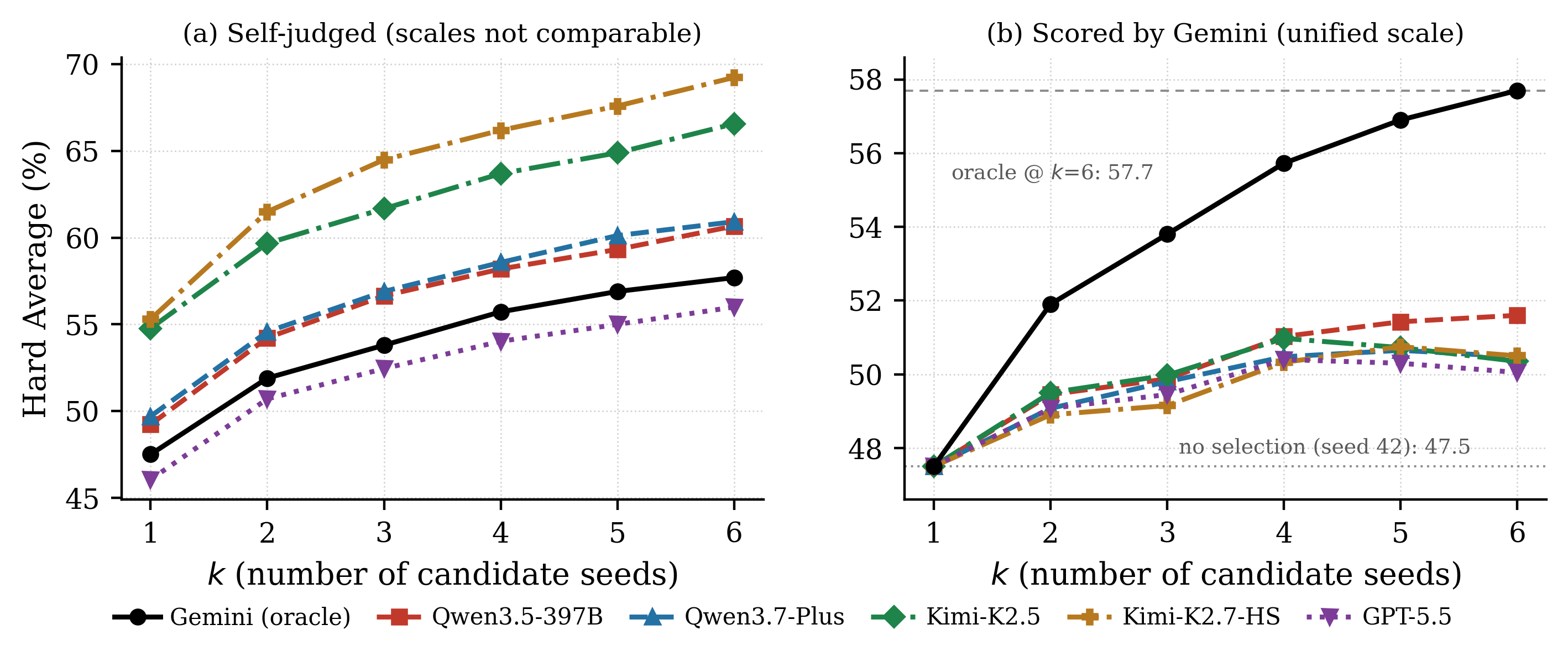}
   \caption{Best-of-$K$ selection on BizGenEval over 400 items on the hard track. (a) Each verifier scores its own pick. (b) The same picks rescored by Gemini-3-Flash-Preview. Hard average values are mean per-item checklist accuracy on the hard track, whcih differ from error score (Appendix~\ref{app:benchmarks}).}
   \label{fig:bestofk}
\end{figure*}

Table~\ref{tab:wise_results} shows that \textsc{Stride} achieves the best overall WiScore.
\textsc{Stride} (ERNIE-Image) obtains 0.92 and \textsc{Stride} (Qwen-Image-2512) obtains 0.91, outperforming the strongest prior agentic baseline, GenEvolve~\citep{chen2026genevolve} (0.82), by 0.10 and 0.09, respectively.
They also exceed the strongest proprietary model, Nano-Banana-Pro~\citep{googledeepmindGemini3ProImageModelCard} (0.87), despite using open-weight generators with an all-open-weight agent stack.
The gains are consistent across domains: \textsc{Stride} (ERNIE-Image) scores at least 0.87 in every category and achieves the best result in Time, Biology, Physics, and Chemistry, improving ERNIE-Image’s standalone score from 0.58 to 0.92.
And \textsc{Stride} (Qwen-Image-2512) leads in Culture and Space.

These results suggest that \textsc{Stride}'s benefit is not limited to checklist-based infographic evaluation.
By dynamically instantiating rubrics from prompts, the same scale-aware refinement loop also improves knowledge-grounded generation where the target entity is only indirectly specified.
Figure~\ref{fig:more_qualitative_results_wise} compares \textsc{Stride} with contemporary agentic methods. \textsc{Stride} produces more consistent, realistic, and aesthetically appealing results, further illustrating the framework's scalability.

\section{Human-MLLM Studies Setup}
\label{app:human_mllm_setup}

\textbf{Study Objectives and Sampling.}
This study assesses two aspects of our evaluation: agreement between
automated evaluators and human judgments, and whether the improvement
of \textsc{Stride} over its frozen generator is preserved under human
evaluation.
As summarized in Table~\ref{tab:human_mllm_protocol}, we sample 25
prompts for each combination of benchmark
(BizGenEval or IGenBench) and generator
(ERNIE-Image or Infographic-V2).
Each prompt contributes two images: the frozen-generator output and
the final \textsc{Stride} output.
The study therefore contains 100 prompt--generator pairs and 200
images. Pairing controls for prompt identity within each generator.

\begin{table}[!htbp]
   \centering
   \footnotesize
   \setlength{\tabcolsep}{4.2pt}
   \caption{Best-of-$6$ selection on BizGenEval over 400 items, reported as mean checklist accuracy on the hard track. Self is the verifier's score of its own pick. Gemini is that pick rescored by Gemini. Capture is the fraction of the Gemini gain from $47.5$ to $57.7$ that each verifier realizes. Leniency is Self minus Gemini at $k=1$, where the images coincide. Inflation is Self minus Gemini at $k=6$.}
   \label{tab:bok_selector}
   \begin{tabular}{@{}l c c c c c@{}}
      \toprule
      \textbf{Selector} & \textbf{Self} & \textbf{Gemini} & \textbf{Capture} & \textbf{Leniency} & \textbf{Inflation} \\
      \midrule
      Gemini (Oracle)   & 57.7          & 57.7            & 100\%            & $+0.0$            & $+0.0$             \\
      \rowcolor{OursPink}
      Qwen3.5-397B-A17B & 60.7          & 51.6            & 40.2\%           & $+1.7$            & $+9.1$             \\
      Qwen3.7-Plus      & 60.9          & 50.5            & 29.2\%           & $+2.2$            & $+10.4$            \\
      Kimi-K2.7-HS      & 69.3          & 50.5            & 29.4\%           & $+7.8$            & $+18.8$            \\
      Kimi-K2.5         & 66.6          & 50.4            & 27.9\%           & $+7.3$            & $+16.2$            \\
      GPT-5.5           & 56.0          & 50.1            & 25.0\%           & $-1.4$            & $+5.9$             \\
      \bottomrule
   \end{tabular}
\end{table}

\textbf{Evaluation Sources and Rubrics.}
Each image is evaluated using human judgments, the default
Qwen3.5-397B-A17B verifier, and the corresponding official benchmark
evaluator: Gemini-3-Flash-Preview for BizGenEval and Gemini-2.5-Pro
for IGenBench.
All three evaluation sources assess the same atomic questions for
each image, and take the average of multiple evaluations.
For BizGenEval, the human study uses the 10 released hard-track
questions. For IGenBench, it uses all released atomic questions
associated with the sampled prompt.
Human evaluation is thus rubric-based rather than a holistic
aesthetic preference assessment.

\section{Diagnostic Experiments}

\subsection{Choice of Rubric-Based Verifier}

To identify a suitable rubric-based verifier, we compare verifier candidates via best-of-$k$ selection on all 400 BizGenEval items, using nested pools of $k\in\{1,\ldots,6\}$ ERNIE-Image samples, totaling 2,400 images.
Each verifier ranks the $k$ images by its assigned mean checklist accuracy (Appendix~\ref{app:benchmarks}) and selects the top-ranked image.
Figure~\ref{fig:bestofk}(a) reports self-judged performance, where each verifier both selects and scores its own best-of-$k$ image, while Figure~\ref{fig:bestofk}(b) reports Gemini-scored performance, where Gemini-3-Flash-Preview re-evaluates the selected images across verifiers.
For each non-Gemini judge, the gap between its self-score and Gemini rescore increases with $k$.

As shown in Table~\ref{tab:bok_selector}, Qwen3.5-397B-A17B achieves a Gemini-scored average checklist accuracy of $51.6$ at $k=6$ and captures $40.2\%$ of Gemini's selection gain, outperforming even its proprietary counterpart, Qwen3.7-Plus~\citep{teamQwen37PlusMultimodalAgent2026}.
Other open-weight judges and GPT-5.5~\citep{openaiGPT55SystemCard} capture only $25.0\%$ to $29.4\%$, whereas Qwen3.5-397B-A17B maintains a capture rate near $40\%$ from $k=2$ to $k=6$.
To avoid rubric-agnostic or overly permissive judging, as observed for Kimi-K2.5~\citep{team2026kimi} and Kimi-K2.7-HS~\citep{moonshotaiKimiK27Code}, and based on comparisons between open-weight and proprietary models, we adopt the open-weight Qwen3.5-397B-A17B as our default rubric-based verifier.

\subsection{Choice of Perceptor}
\label{app:perception_comparison}

The perceptor is an important but replaceable component in \textsc{Stride}. Its role is not to decide whether a candidate should be admitted, as champion selection is still governed by the rubric-based verifier score. Instead, the perceptor provides open-form visual grounding for the lead editor and the retoucher, including visible text blocks, their bounding boxes, and whether the rendered text appears corrupted. This distinction matters because the retoucher can only repair a failure that has been localized. The choice of perception backbone therefore bounds the practical effectiveness of local editing, but it is not a fixed assumption of the framework.

\textbf{Protocol.}
We select the default perceptor backbone through a small, stratified agreement study and practical closed-loop verification. The study uses 20 generated infographics sampled across the five BizGenEval domains and two base generators, ERNIE-Image and Infographic-V2. Each candidate perceptor backend is asked to return visible text blocks and normalized bounding boxes. Since generated images do not come with ground-truth text-box annotations, we use Gemini-3-Flash-Preview as a proxy reference after sanity-checking its localization behavior against a dedicated PP-OCRv5 server~\citep{cui2026pp}. PP-OCRv5 server is therefore included in the table as a cross-check/reference OCR baseline, rather than as the perception backbone selected for \textsc{Stride}. We report \emph{agreement} with the reference, not absolute accuracy.

\begin{figure*}[!htbp]
   \centering
   \includegraphics[width=\textwidth]{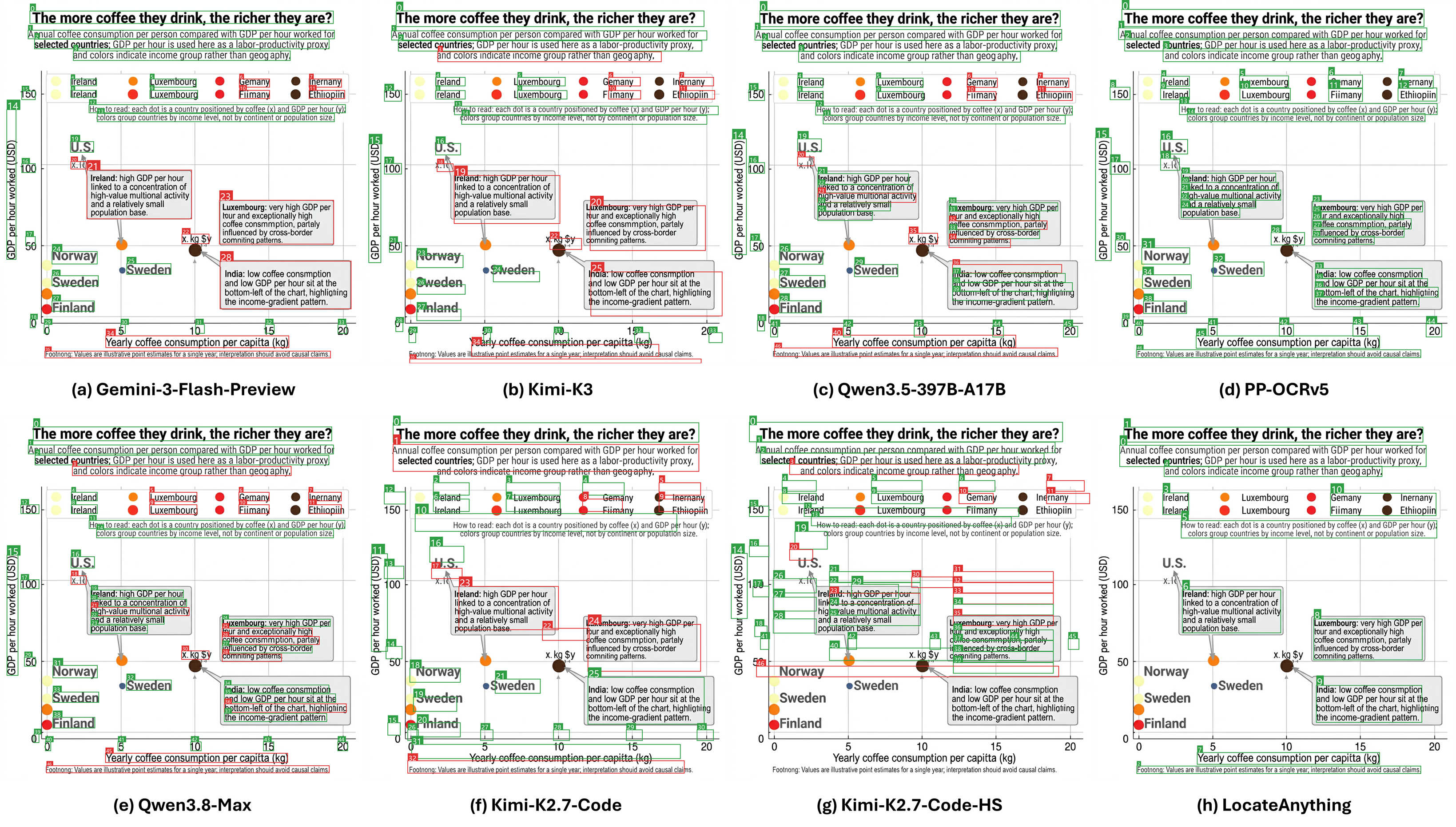}
   \caption{Qualitative comparison of perceptor outputs on generated infographics.
      Different perception backbones identify visible text blocks, localize them with bounding boxes, and assess whether the rendered text is garbled.
      Green boxes indicate regions that the perceptor identifies as correct, whereas red boxes indicate regions it identifies as incorrect.
      The comparison highlights the trade-off between geometric localization and contextual text understanding that motivates the use of an MLLM-based perceptor in \textsc{Stride}.}
   \label{fig:viz_perceptor_comparison}
\end{figure*}

\begin{table}[t]
   \centering
   \small
   \setlength{\tabcolsep}{3pt}
   \caption{Perception-backbone agreement with Gemini-3-Flash-Preview on 20 stratified generated infographics under a
      box-first protocol. PP-OCRv5 is reported as a cross-check/reference OCR baseline. Metrics measure agreement with
      the proxy reference, not absolute accuracy. Higher is better for A-R, A-P, A-F1, mean IoU, NED-sim, and exact;
      lower is better for anchor offset and latency.}
   \label{tab:perception_agreement}
   \resizebox{\linewidth}{!}{%
      \begin{tabular}{lcccccccc}
         \toprule
         \textbf{Model}    & \textbf{A-R}               & \textbf{A-P}           & \textbf{A-F1}     & \textbf{mean IoU} & \textbf{NED-sim} &
         \textbf{exact}    & \textbf{Anchor med. / P90} & \textbf{Med. latency}                                                               \\
         \midrule
         PP-OCRv5          & \textbf{0.881}             & \textbf{0.703}         & \textbf{0.764}    & 0.729             & 0.899            &
         0.590             & \textbf{2.2 / 6.6}         & 44s                                                                                 \\
         \rowcolor{OursPink}
         Kimi-K3           & \underline{0.738}          & \underline{0.699}      & \underline{0.713} & 0.730             &
         \underline{0.949} & 0.721                      & \underline{4.6 / 16.3} & 49s                                                        \\
         Qwen3.5-397B-A17B & 0.511                      & 0.486                  & 0.495             & \underline{0.749} & \textbf{0.957}   &
         \textbf{0.751}    & 6.0 / 46.3                 & 71s                                                                                 \\
         Qwen3.8-Max       & 0.388                      & 0.360                  & 0.370             & 0.670             & 0.924            &
         \underline{0.728} & 13.4 / 148.4               & 259s                                                                                \\
         LocateAnything-3B & 0.345                      & 0.535                  & 0.322             & \textbf{0.777}    & --               &
         --                & --                         & \textbf{6s}                                                                         \\
         Kimi-K2.7-Code-HS & 0.318                      & 0.292                  & 0.300             & 0.654             & 0.896            &
         0.698             & 16.0 / 98.6                & \underline{29s}                                                                     \\
         Kimi-K2.7-Code    & 0.283                      & 0.273                  & 0.273             & 0.655             & 0.869            &
         0.688             & 19.1 / 92.0                & 234s                                                                                \\
         \bottomrule
      \end{tabular}%
   }
\end{table}

We use a box-first protocol. Candidate boxes are first matched to reference boxes by IoU, and the text metrics are computed only on matched pairs. Unmatched reference boxes reduce agreement recall (A-R), while unmatched candidate boxes reduce agreement precision (A-P). A-F1 is their harmonic mean. Mean IoU measures the spatial overlap of matched boxes. NED-sim is one minus normalized edit distance between the matched transcriptions, and exact is the fraction of matched pairs with identical text. We also report an anchor-offset diagnostic: for text strings that appear uniquely in both outputs, we measure the distance between the two box centers in the normalized 0--1000 coordinate system. Its median measures typical localization quality, while its P90 exposes long-tail failures.
We provide a qualitative comparison of different perceptors in Figure~\ref{fig:viz_perceptor_comparison}.

\begin{table*}[!htbp]
   \centering
   \small
   \setlength{\tabcolsep}{3.2pt}
   \renewcommand{\arraystretch}{1.08}
   \caption{Comparison of champion-evolution strategies on BizGenEval.
      Each entry reports the official error score as
      hard (easy).
      All non-oracle variants use the same refinement framework and differ only
      in how new champion candidates are selected and admitted.
      The $(1,1)$ and $(1,4)$ variants denote forced replacement, whereas
      the $(1+1)$ and $(1+4)$ variants retain the incumbent champion unless
      a better candidate is found.
      Oracle admission keeps Qwen3.5-397B-A17B within-round candidate selection
      unchanged but uses Gemini-3-Flash-Preview for admission; Full oracle uses Gemini-3-Flash-Preview for
      both candidate selection and admission.}
   \label{tab:evolution_strategy}
   \resizebox{\textwidth}{!}{%
      \begin{tabular}{@{}l|ccccc|cccc|c@{}}
         \toprule
         \multirow{2}{*}{\textbf{Strategy}}
          & \multicolumn{5}{c|}{\textbf{Content Domain}}
          & \multicolumn{4}{c|}{\textbf{Capability}}
          & \multirow{2}{*}{\textbf{Average}}                \\
         \cmidrule(lr){2-6}
         \cmidrule(lr){7-10}
          & Slides
          & Webpage
          & Poster
          & Chart
          & Sci.\ Fig.
          & Layout
          & Attr.
          & Text
          & Know.
          &                                                  \\
         \midrule

         \rowcolor{GroupGray}
         \multicolumn{11}{l}{\textit{Non-oracle strategies}} \\

         Random selection
          & 45.0 (74.5)
          & 59.8 (77.2)
          & 47.5 (\underline{73.8})
          & 21.3 (50.0)
          & 28.5 (67.0)
          & 54.6 (78.2)
          & 40.0 (77.6)
          & 46.4 (\underline{77.0})
          & 20.6 (41.2)
          & 40.4 (68.5)
         \\

         $(1,1)$ forced replacement
          & 47.5 (71.8)
          & 55.0 (77.2)
          & 44.2 (65.0)
          & 22.0 (43.5)
          & 28.0 (60.8)
          & 56.0 (75.2)
          & 37.8 (73.8)
          & 42.2 (72.8)
          & 21.4 (32.8)
          & 39.4 (63.7)
         \\

         $(1+1)$-evolution strategy
          & 53.0 (75.0)
          & \underline{60.2} (81.2)
          & \underline{50.3} (72.5)
          & \underline{29.8} (\underline{53.5})
          & \underline{34.5} (65.5)
          & \underline{62.6} (\underline{80.0})
          & \underline{43.6} (\underline{81.0})
          & 49.2 (75.8)
          & 26.8 (41.4)
          & \underline{45.6} (\underline{69.5})
         \\

         $(1,4)$ forced replacement
          & \underline{56.0} (\underline{76.2})
          & \textbf{62.0} (\underline{81.5})
          & 49.3 (72.2)
          & 27.8 (49.8)
          & 32.0 (\underline{67.2})
          & 59.2 (77.2)
          & 42.4 (77.8)
          & \underline{50.4} (\textbf{79.8})
          & \underline{29.6} (\underline{42.8})
          & 45.4 (69.4)
         \\

         \rowcolor{OursPink}
         \textbf{$(1+4)$-evolution strategy (ours)}
          & \textbf{61.3} (\textbf{79.2})
          & \underline{60.2} (\textbf{83.2})
          & \textbf{54.5} (\textbf{74.5})
          & \textbf{37.0} (\textbf{62.3})
          & \textbf{42.5} (\textbf{69.2})
          & \textbf{64.4} (\textbf{80.8})
          & \textbf{50.8} (\textbf{82.6})
          & \textbf{55.2} (\textbf{79.8})
          & \textbf{34.0} (\textbf{51.6})
          & \textbf{51.1 (73.7)}
         \\

         \midrule
         \rowcolor{GroupGray}
         \multicolumn{11}{l}{\textit{Oracle diagnostics}}    \\

         Oracle admission (Qwen + Gemini)
          & 63.0 (79.8)
          & 70.5 (85.0)
          & 60.5 (74.0)
          & 42.3 (60.8)
          & 48.3 (74.0)
          & 68.2 (80.8)
          & 56.8 (83.0)
          & 61.0 (81.4)
          & 41.6 (53.6)
          & 56.9 (74.7)
         \\

         Full oracle (Gemini + Gemini)
          & 70.2 (87.8)
          & 76.0 (93.2)
          & 70.5 (86.8)
          & 47.8 (68.0)
          & 57.5 (82.5)
          & 71.8 (87.4)
          & 63.6 (86.2)
          & 67.2 (88.2)
          & 54.8 (72.8)
          & 64.4 (83.7)
         \\

         \bottomrule
      \end{tabular}%
   }
\end{table*}

\textbf{Agreement Results.}
Table~\ref{tab:perception_agreement} shows that no single backend dominates every criterion. PP-OCRv5 has the strongest box coverage and the smallest anchor offsets, which confirms that the cross-check OCR baseline is a strong text localizer. However, its transcription agreement is substantially weaker than the strongest MLLM backends: its exact-match rate is 0.590, compared with 0.721 for Kimi-K3~\citep{team2026kimik3} and 0.751 for Qwen3.5-397B-A17B~\citep{teamQwen35NativeMultimodal2026}. Qwen3.5-397B-A17B achieves the best matched-text similarity, but its recall and precision are much lower and its P90 anchor offset is nearly three times that of Kimi-K3, indicating a heavier tail of localization failures. Kimi-K3 provides the most balanced open-weight tradeoff: its A-F1 is close to the OCR cross-check baseline PP-OCRv5, its mean IoU is comparable, and its transcription agreement is much higher than the conventional OCR baseline, including PP-OCRv5 and LocateAnything~\citep{wang2026locateanything}.

\textbf{Why OCR Alone is Insufficient.}
The perceptor in \textsc{Stride} is not a generic OCR module. It must supply spatial evidence that is useful for deciding whether a failed rubric can be repaired locally. Conventional OCR and detection models~\citep{zhou2024yolo,cui2026pp,wang2026locateanything} are optimized to find text-like regions and transcribe them, but generated infographics often contain pseudo-text artifacts: glyphs that look like text locally but are misspelled, structurally broken, or semantically nonsensical in context. In such cases, as shown in Figure~\ref{fig:viz_perceptor_comparison}, a pure OCR system (like PP-OCRv5 in Figure~\ref{fig:viz_perceptor_comparison} (d) or LocateAnything in Figure~\ref{fig:viz_perceptor_comparison} (h)) can return precise boxes around corrupted glyphs and treat them as normal readable text. These false-positive text blocks are harmful for the retoucher, because they create repair targets that appear well localized while failing to represent reliable semantic evidence. An MLLM perceptor is better aligned with this use case because it jointly considers the box, the transcription, and the surrounding linguistic and visual context, including whether the text should be marked as garbled.

\textbf{Backbone Choice.}
We choose Kimi-K3 as the default perceptor for reasons beyond single-call localization agreement. First, it offers the best overall balance between localization coverage and text understanding among the open-weight candidates in Table~\ref{tab:perception_agreement}. Second, the rubric-based verifier in our default stack uses a Qwen-series model. Reusing the same model family for perception would increase the risk of correlated errors between the grounding signal and the verification signal, weakening the benefit of the dual-route design. Third, \textsc{Stride} does not rely on a single perception sample. Each refinement round aggregates 4 to 10 OCR/perception samples, and rollback triggers resampling in later rounds. This repeated-sampling design directly mitigates the long-tail localization failures observed in single-call diagnostics.

\subsection{Choice of Evolution Strategy}
\label{app:evolution_strategy}

\textbf{Evolution Strategy Experiment Setup.} Table~\ref{tab:evolution_strategy} compares alternative
champion-evolution strategies on BizGenEval. Each entry reports the
official error score as hard (easy), with higher values
indicating better performance. We use the Hard track as the primary
evaluation protocol and report Easy-track scores for reference.
The comparison separates two decisions in champion evolution. First,
the \emph{pick} decision selects the best candidate within the current new
champion candidate set. Second, the \emph{admission} decision determines whether
the picked candidate should replace the incumbent champion. In the
default $(1+4)$-evolution strategy, both decisions are made by the Qwen3.5-397B-A17B
verifier. The oracle-admission variant keeps Qwen3.5-397B-A17B for within-round
candidate selection but uses the Gemini-3-Flash-Preview evaluator for
admission. The full-oracle variant uses Gemini-3-Flash-Preview for both candidate
selection and admission.

\textbf{Champion Candidates.}
Among the non-oracle variants, our default $(1+4)$-evolution strategy
achieves the highest average hard score, reaching 51.1. Increasing the
candidate count improves both forced-replacement and evolution strategies:
the forced-replacement score increases from 39.4 for $(1,1)$ to 45.4
for $(1,4)$, while the evolution strategy score increases from 45.6 for $(1+1)$ to
51.1 for $(1+4)$. Thus, generating multiple proposals provides a
substantial search benefit under both admission rules.

\textbf{Incumbent Retention.}
Incumbent retention provides an additional and independent benefit.
With one candidate, the evolution strategy improves the hard average from
39.4 for forced replacement $(1,1)$ to 45.6 for $(1+1)$, a gain of
6.2 points. With four candidates, the corresponding improvement is
5.7 points, from 45.4 for forced replacement $(1,4)$ to 51.1 for
$(1+4)$. Therefore, the performance of the proposed strategy is not
explained by increased candidate sampling alone: it also depends on
rejecting non-improving updates and preserving a strong incumbent.

\textbf{Selection Methods.}
Random selection reaches an average hard score of 40.4, which is
substantially below the proposed $(1+4)$ strategy. It slightly
outperforms forced replacement $(1,1)$, which reaches 39.4, indicating
that random replacement can occasionally benefit from stochastic
diversity. However, it remains well below the strategies that combine
multiple proposals with score-guided selection and incumbent retention.
This result suggests that repeated stochastic modification alone is
insufficient for reliable refinement.

\textbf{Oracle Experiments in Evolution Strategy.}
We also conducted oracle evaluations in our evolution-strategy comparison.
In Table~\ref{tab:evolution_strategy}, replacing only the Qwen-based admission decision with the
Gemini-based benchmark evaluator increases the Hard average from 51.1
to 56.9, a gain of 5.8 points, while the corresponding Easy score
increases from 73.7 to 74.7. Because within-round candidate selection
remains Qwen-based in this setting, this improvement isolates the
effect of the admission decision. The result indicates that the
default verifier sometimes accepts inferior updates or rejects
beneficial ones, with the largest Hard-score gains appearing in
Knowledge (+7.6), Attribute (+6.0), and Text (+5.8).

The full-oracle variant, which uses Gemini for both within-round
candidate selection and admission, further increases the Hard average
to 64.4. This is 7.5 points above oracle admission and 13.3 points
above the default $(1+4)$ strategy. Since admission is already
controlled by Gemini in both oracle variants, the additional gap
isolates errors in Qwen's within-round candidate ranking. The largest
additional gain is observed for Knowledge, which improves from 41.6
under oracle admission to 54.8 under the full oracle. These results
show that both candidate ranking and incumbent admission contribute
to the remaining performance gap.

Overall, the comparison supports the $(1+\lambda)$-evolution strategy as
a central component of \textsc{Stride}. Multiple candidates improve the
search space explored in each refinement round, while incumbent
retention prevents the system from replacing a strong champion with a
lower-quality proposal. The oracle results further show that the
effectiveness of this mechanism is bounded by the quality of the
verifier used for candidate ranking and admission.

\section{More Qualitative Results}
\label{app:more_qualitative_results}

\begin{figure*}[!htbp]
   \centering
   \includegraphics[width=\textwidth]{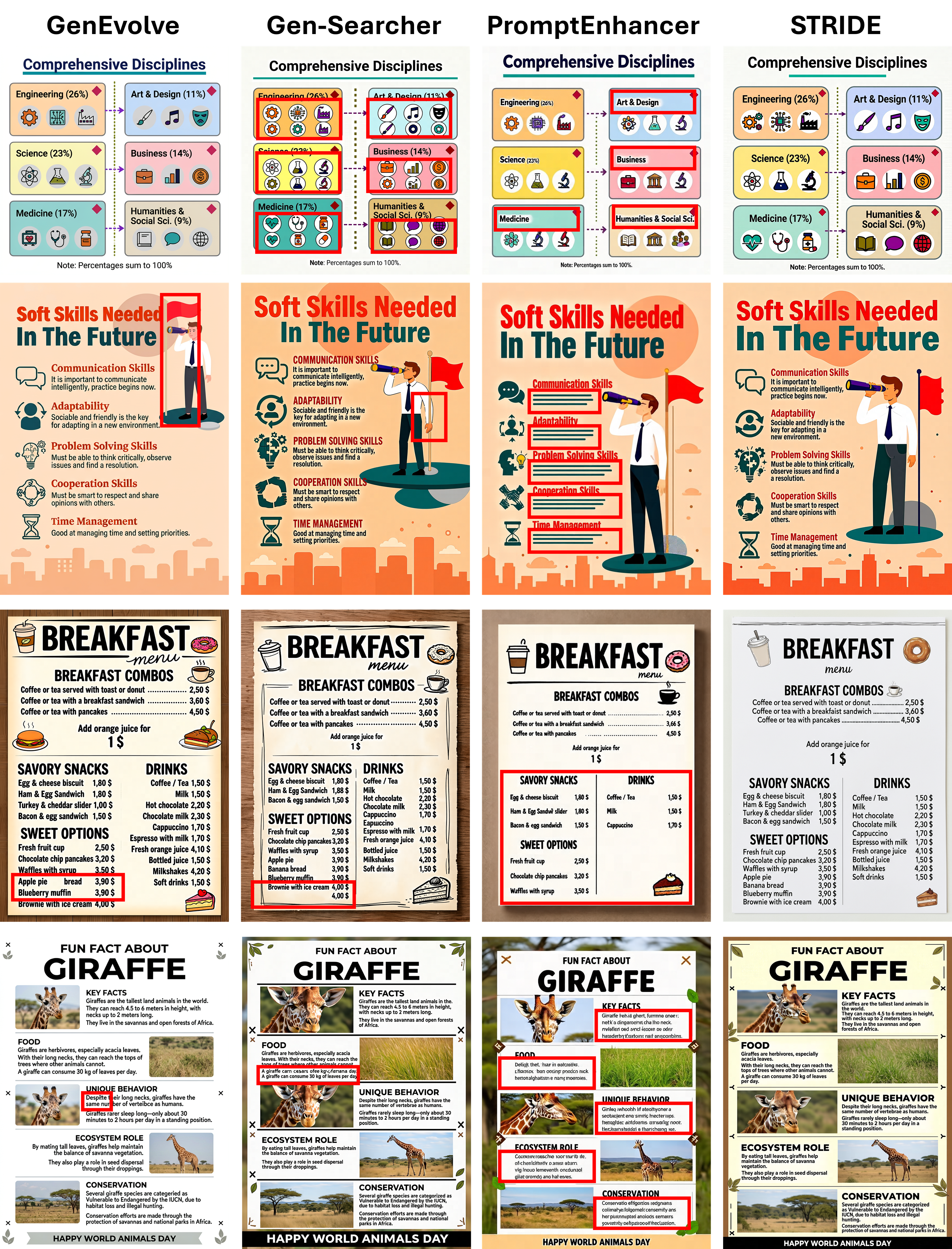}
   \caption{More qualitative comparison with existing agentic methods on the infographic task.}
   \label{fig:more_qualitative_results}
\end{figure*}

\begin{figure*}[!htbp]
   \centering
   \includegraphics[width=\textwidth]{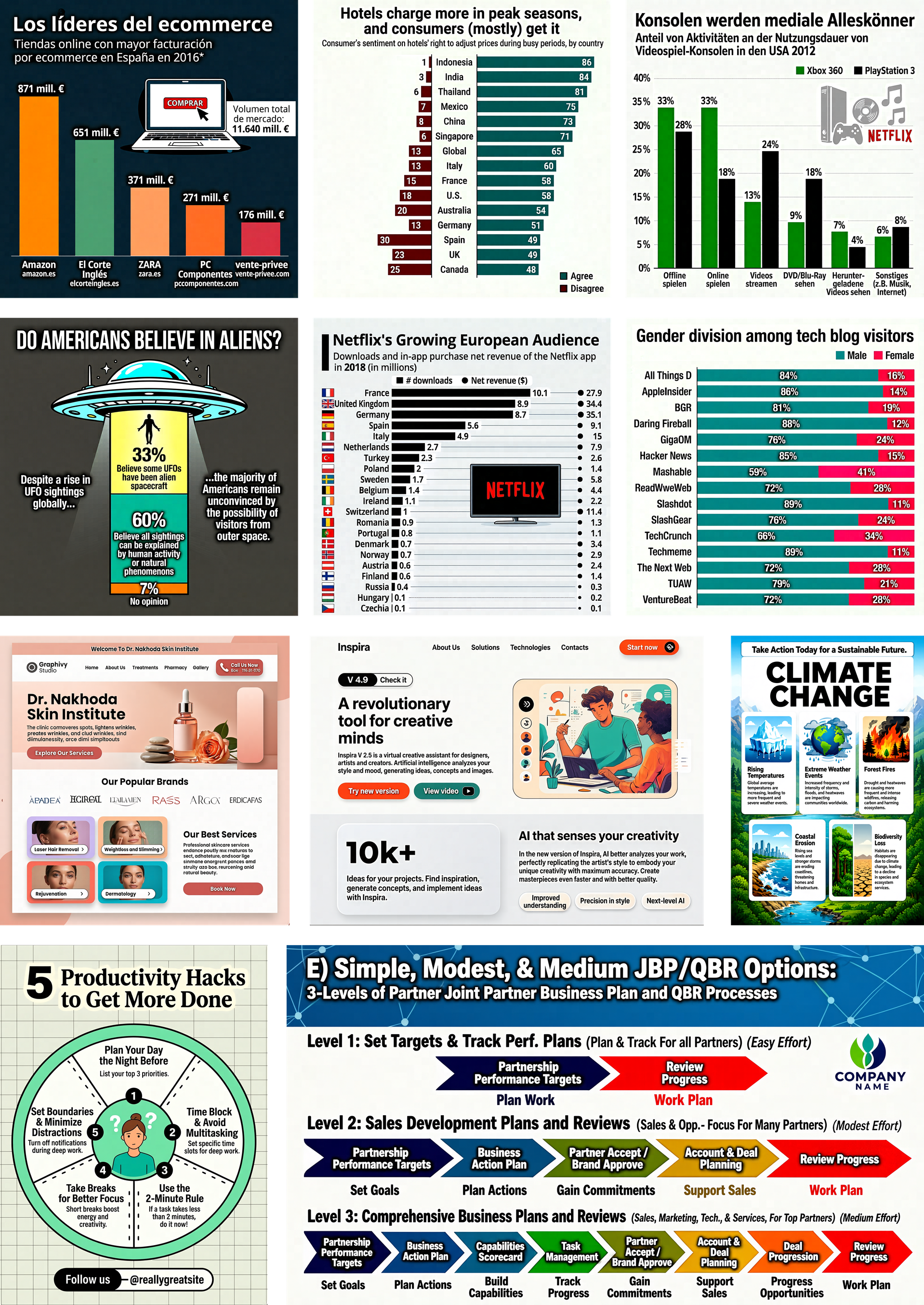}
   \caption{More qualitative results of \textsc{Stride}.}
   \label{fig:more_qualitative_results_stride}
\end{figure*}

Figures~\ref{fig:more_qualitative_results} and~\ref{fig:more_qualitative_results_stride}
present additional qualitative results on infographic generation. Across the shown examples, \textsc{Stride} produces more complete and coherent infographics than existing agentic methods, with better preservation of the requested layout, attributes, text, and semantic content. In contrast, the compared methods more frequently exhibit missing or distorted text, incorrect element arrangements, and unresolved prompt constraints. These examples further illustrate the benefit of \textsc{Stride}'s rubric-guided diagnosis, scale-aware execution, and iterative champion refinement.

\begin{figure*}[!htbp]
   \centering
   \includegraphics[width=\textwidth]{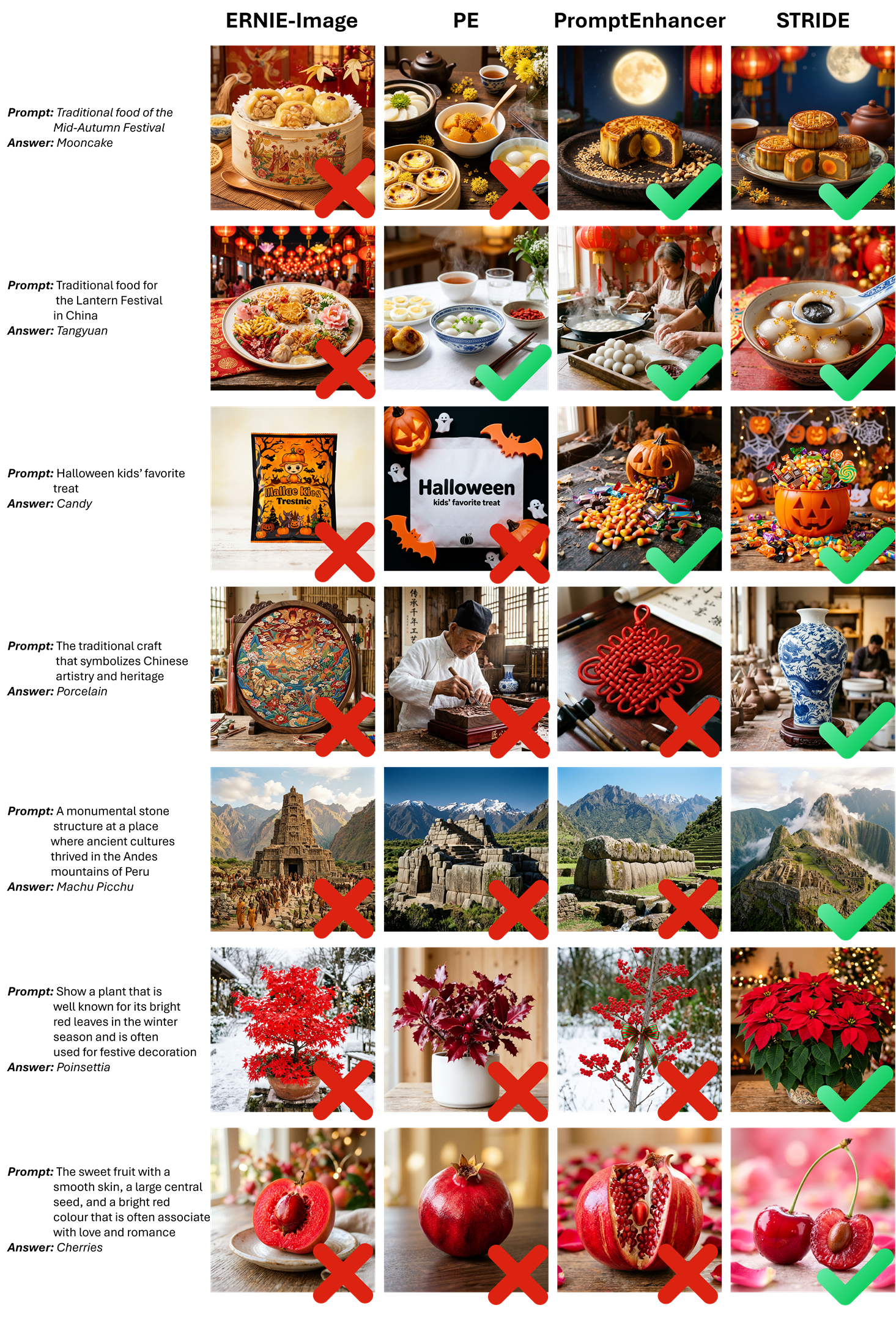}
   \caption{Qualitative comparison with existing agentic methods on general WISE task.}
   \label{fig:more_qualitative_results_wise}
\end{figure*}

Figure~\ref{fig:more_qualitative_results_wise} further evaluates the transferability of \textsc{Stride} on the general WISE task. Although WISE is not restricted to infographic-specific requirements, \textsc{Stride} generates results that more faithfully preserve the prompt-specified objects, relationships, and visual structure than the compared agentic methods. The consistent improvements across these examples suggest that \textsc{Stride}'s refinement mechanism is not tied to a particular visual domain and can generalize to tasks with prompt-derived requirements.

\section{Case Study of Stride}
\label{app:case_study}

\begin{figure*}[!htbp]
   \centering
   \includegraphics[width=\textwidth]{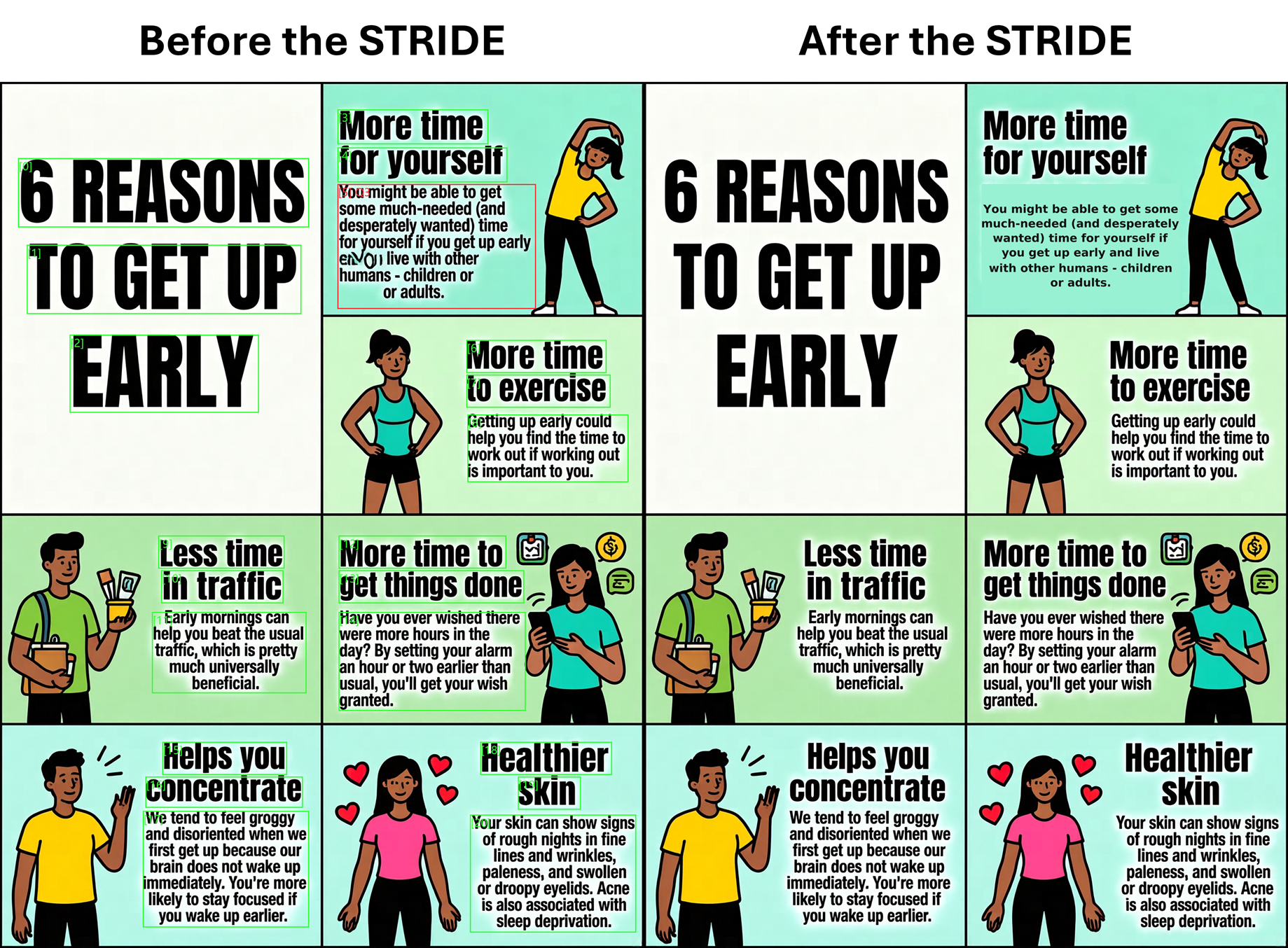}
   \caption{Case study 1: Localized text repair with \textsc{Stride}. Guided by visual grounding, the lead editor selects local editing to correct a garbled text block while preserving the surrounding content and layout. Green and red boxes mark regions identified as correct and erroneous, respectively.}
   \label{fig:case_study_results_1}
\end{figure*}

\begin{figure*}[!htbp]
   \centering
   \includegraphics[width=\textwidth]{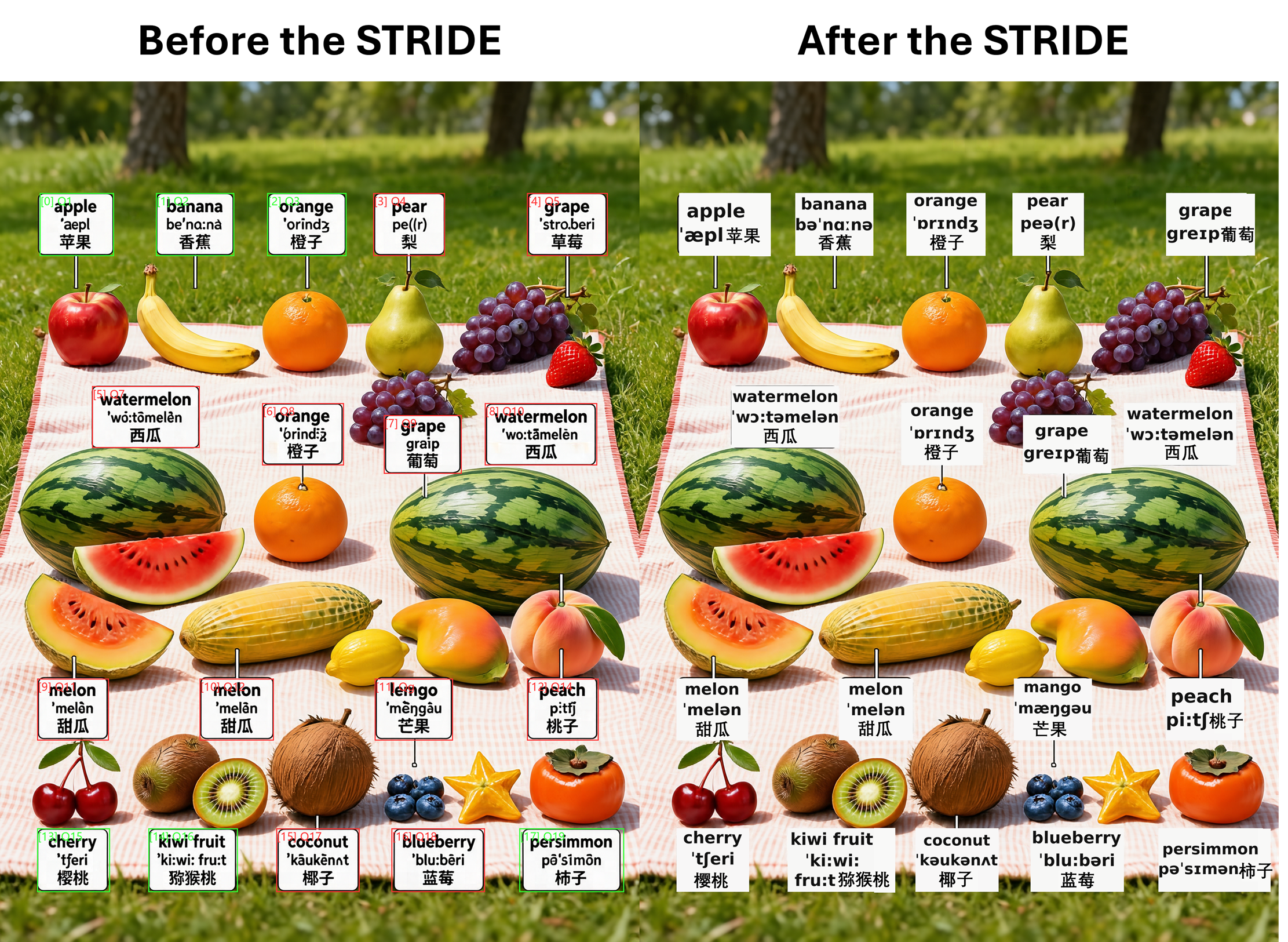}
   \caption{Case study 2: Multiple text corrections with \textsc{Stride}. Despite low initial verification scores, the lead editor identifies localized label errors and selects local editing to correct phonetic transcriptions and a Chinese character while preserving the overall layout. Green and red boxes mark regions identified as correct and erroneous, respectively.}
   \label{fig:case_study_results_2}
\end{figure*}

To illustrate the execution process of \textsc{Stride}, we present two representative cases in Figures~\ref{fig:case_study_results_1} and~\ref{fig:case_study_results_2}. Each figure shows the initial champion, the perceptor's visual grounding, and the refined result, where green and red boxes indicate correctly and incorrectly satisfied regions, respectively. We additionally report the lead editor's intermediate analysis to make the diagnosis and route-selection process explicit.

\textbf{Case 1: Localized text repair.}
In Figure~\ref{fig:case_study_results_1}, the initial champion satisfies most of the requirements and has a coherent global layout. The perceptor identifies the only remaining failure as a garbled text block, while reporting healthy layout coverage and no region overlap. Based on this evidence, the lead editor correctly classifies the failure as a localized text-rendering error and selects the surgical editing route rather than global redraw. The retoucher therefore modifies only the erroneous text region. The refined result resolves the detected text error while preserving the surrounding layout and already-correct content, demonstrating that \textsc{Stride} can perform conservative, targeted refinement when the global composition is already valid.

\begin{promptbox}[Lead Editor Analysis Result]{CigarBlue}{CigarBlueBg}
   {
      'reasoning': 'The subject appears in OCR blocks (garbled block). Text problem -> surgical. High accuracy, healthy
      layout metrics. Fix the one garbled block.',
      'ok': True,
      'kind': 'analysis',
      'reason': 'agent: The only failure is Q3, whose paragraph is present in the OCR blocks but garbled, while layout
      metrics (coverage 38.6%, overlap 0%, hard_acc 1.00) are healthy, so an in-place text fix suffices.',
      'suggest': 'surgical',
      'model': 'kimi-k3',
      'failure_mode': 'garbled_text_block',
      'confidence': 0.9
   }
\end{promptbox}

\textbf{Case 2: Multiple localized text errors.}
Figure~\ref{fig:case_study_results_2} presents a more challenging initial champion with a low verification accuracy of $0.10$. Although nine hard and nine easy questions initially fail, the perceptor finds that all relevant subjects are present in the OCR blocks and that the layout has no region overlap. The main errors are concentrated in garbled phonetic lines and an incorrect Chinese glyph, rather than missing objects or structural layout failures. The lead editor consequently selects local text repair instead of redrawing the entire image. After refinement, the erroneous labels are corrected while the overall arrangement and visual structure remain intact. This case shows that \textsc{Stride} does not equate a low score with a global failure; it uses grounded evidence to distinguish localized rendering errors from structural errors and to select the least disruptive effective action.

Together, these cases demonstrate the reliability of \textsc{Stride}'s refinement loop: the perceptor turns abstract verification failures into spatially actionable evidence, the lead editor uses this evidence together with the agent state to select an appropriate editing scale, and the retoucher performs targeted corrections while preserving valid content.

\begin{promptbox}[Lead Editor Analysis Result]{CigarBlue}{CigarBlueBg}
   {
      'reasoning': 'Let me analyze this carefully. Current state: accuracy=0.10, hard_acc=0.10, easy_acc=0.10;
      n_failed_hard=9, n_failed_easy=9; OCR has 18 blocks, 12 garbled; box_coverage=17.1%; box_overlap_rate=0.0%;
      champion_changed_since_last_round=true; first attempt. All failed subjects appear in OCR text blocks, but many are
      garbled, especially the phonetic second line. This makes the issue surgical text repair rather than redraw..... .',
      'ok': True,
      'kind': 'analysis',
      'reason': "agent: Layout is intact (0% overlap, all 18 fruit labels present in OCR) and every failed question's
      subject is visible but has a wrong IPA middle line or one wrong Chinese glyph, so this is in-place text correction,
      not a redraw.",
      'suggest': 'surgical',
      'model': 'kimi-k3',
      'failure_mode': 'garbled_ipa_label_text',
      'confidence': 0.78
   }
\end{promptbox}

\section{Limitations and Future Work}

\textsc{Stride} has two limitations that suggest directions for future work. First, although the open-weight configuration achieves competitive results, refinement remains dependent on the reliability of rubric verification and spatial grounding. Monotonic admission applies to the internal verifier score and does not guarantee improvement under independent evaluation. Our diagnostic experiments indicate room for improvement in these components, which the modular framework can readily accommodate. Better-calibrated verification and more precise grounding could further strengthen candidate selection and targeted correction.

Second, the current retoucher focuses on text and formula replacement within localized regions, enabling targeted corrections while preserving the surrounding content. Changes to chart geometry or spatial relationships currently rely on the composer’s global regeneration. Extending local refinement to structured visual elements would allow a broader range of errors to be corrected without redrawing the entire image. A promising direction is to jointly edit labels and their associated visual marks, further extending the benefits of scale-aware refinement.

\section{Prompts for Agents in Stride}
\label{app:agent-prompts}

To facilitate reproduction and inspire future research, this section reports the prompts used by the agents in \textsc{Stride}. Variables enclosed by braces are filled at inference time.

\subsection{Verifier}
\label{app:agent-prompts-verifier}

\begin{promptbox}[Verifier System Prompt]{CigarBlue}{CigarBlueBg}
   You are a precise, strict visual checklist judge.
\end{promptbox}

\begin{promptbox}[Verifier Prompt]{CigarBlue}{CigarBlueBg}
   Evaluate the image against EACH checklist question below. For each question answer strictly true (the image fully satisfies it) or false (it does not, or only partially). Be strict: partial credit = false.

   === CHECKLIST ({n_questions} questions) ===
   {qlines}

   Return STRICT JSON only: {{"results":[{{"qid":1,
                  "result":true,
                  "reason":"..."}},...]}} covering ALL questions, no extra text.
\end{promptbox}

\subsection{Perceptor}
\label{app:agent-prompts-perceptor}

\begin{promptbox}[Perceptor Prompt]{CigarBlue}{CigarBlueBg}
   You are an OCR + layout localizer. List EVERY visible text block in this image—title, labels, values, formulas, captions, everything. Be thorough: scan the entire image systematically top->bottom, left->right.

   For each text block give:
   1) literal text (verbatim transcription, even if garbled/misspelled)
   2) bounding box [x1,y1,x2,y2] as INTEGERS in 0-1000 range (x=horizontal left->right, y=vertical top->bottom), fit TIGHTLY to the text
   3) garbled flag: true if text has spelling errors, nonsense characters, broken formulas, or looks corrupted.

   IMPORTANT: Do NOT skip any text. List at least {min_blocks}-30 blocks for a typical infographic.

   Return STRICT JSON: {{"blocks":[{{"text":"...",
                  "box":[x1,y1,x2,y2],
                  "garbled":true/false}},...]}} with NO extra commentary.
\end{promptbox}

\subsection{Lead Editor}
\label{app:agent-prompts-lead}

\begin{promptbox}[Lead Editor Prompt]{CigarBlue}{CigarBlueBg}
   Decide how to repair a generated infographic that failed part of its checklist, and write concrete guidance for whichever repair actor you choose.

   === CURRENT STATE ===
   accuracy={accuracy:.2f}  hard_acc={hard_acc:.2f}  easy_acc={easy_acc:.2f}
   n_failed_hard={n_fail_hard}  n_failed_easy={n_fail_easy}

   === OCR VIEW (of the CURRENT champion image) ===
   {ocr_view}

   === OCR TEXT BLOCKS {ocr_blocks}

   === FAILED QUESTIONS ===
   {failed_questions}

   === PRIOR ATTEMPTS ON THIS ITEM ===
   {attempt_history}

   === HISTORICAL PRIORS (guidance, not rules — you may overrule them with reason) ===
   {rules_prior}

   Two strategies are available:
   surgical — OCR the image, compute per-block text corrections, occlude and re-render those blocks in place. Preserves everything already correct. Cheap. Cannot fix layout, composition, or missing visual elements. The renderer repaints text INSIDE the existing bounding box, so it cannot move, resize, add or delete an element.
   redraw   — regenerate the whole image from an enhanced prompt with several seeds. Can fix structure and layout. Discards everything currently correct, so it can regress.

   HOW TO READ THE EVIDENCE
   - A failed question whose subject DOES appear in the OCR text blocks (even garbled) is a TEXT problem -> surgical.
   - A failed question whose subject appears NOWHERE in the OCR text blocks usually means the element was never drawn -> redraw.
   - Low box_coverage or high box_overlap_rate points at a broken layout -> redraw.
   - If champion_changed_since_last_round is false, the OCR numbers are unchanged BY CONSTRUCTION. Do not read "garbled_rate did not drop" as "surgical failed" in that case; use PRIOR ATTEMPTS instead.
   - "BROKE previously-correct" in the history is the strongest signal available: the last repair damaged content that was already right. Say explicitly what to avoid this round.

   YOUR GUIDANCE
   Write `guidance` for the actor you selected. It is injected verbatim into that actor's prompt next round, so address the actor, not the user. Be specific and grounded in the history above: name the blocks, question ids, or constraints involved and say what to do differently. If there is no prior attempt, give a short plan instead. Never invent facts that are not in the evidence above. Keep it under 1000 characters.

   Return STRICT JSON only:
   {{"strategy":"surgical"|"redraw",
         "confidence":0-1,
         "failure_mode":"short_snake_case_label",
         "reason":"one sentence for a human reader",
         "guidance":"concrete instructions for the selected actor"}}
\end{promptbox}

\subsection{Composer}
\label{app:agent-prompts-composer}

\begin{promptbox}[Composer System Prompt]{CigarBlue}{CigarBlueBg}
   You are a prompt engineer for a text-to-image model that generates infographics.

   You are given the ORIGINAL prompt, the FAILED hard checklist questions (the generated image violated these constraints), and the verifier's reasons. You rewrite those failures into short, imperative, unambiguous constraint sentences and APPEND them to the original prompt.

   You are called in up to two passes:
   Pass 1 — no web evidence. Answer from your own knowledge and state your confidence honestly. If the constraints depend on facts you are unsure of, say so (need_search=true) and propose real search queries.
   Pass 2 — only happens if you asked for a search. You are shown your own Pass-1 output plus the search results. Revise the constraints and update confidence. Keeping a low confidence after searching is a valid and useful answer — do not inflate it just to look decisive.

   Rules:
   - NEVER modify, shorten, or reorder the original prompt. Only append.
   - The appended section starts with: "STRICT LAYOUT/COUNT CONSTRAINTS (must all hold):"
   - One numbered line per failed constraint, phrased as a direct instruction (e.g. "Panel (E) must be positioned directly below panel (A).", "Draw EXACTLY five panels labeled (A) (B) (C) (D) (E).", "The legend swatch color for each method must be identical to its bar color.").
   - Keep the appended section under 900 characters. Prioritize count/position constraints.
   - Never invent facts. An honest low confidence is better than a confident wrong constraint, because a wrong constraint actively steers the image model away from the right answer.
   - Output ONLY the JSON object described in the user message. No commentary.
\end{promptbox}

\begin{promptbox}[Composer Prompt]{CigarBlue}{CigarBlueBg}
   ORIGINAL PROMPT (never modify, reorder, or delete any of it — append only):
   {original_prompt}

   === ITEM ===
   domain={domain}  dimension={dimension}

   === CURRENT STATE ===
   accuracy={accuracy:.2f}  hard_acc={hard_acc:.2f}  easy_acc={easy_acc:.2f}
   n_failed_hard={n_fail_hard}

   FAILED HARD QUESTIONS (each must become an explicit, checkable constraint):
   {failed_questions}

   JUDGE REASONS (why the verifier rejected them — the most specific evidence you have):
   {judge_reasons}

   GUIDANCE FROM ANALYZER (based on previous attempts on this image; "(none)" on first try):
   {guidance}

   YOUR PRIOR ANALYSIS (empty on the first pass; on the second pass this is what YOU wrote before seeing any search results — revise it, do not restate it):
   {prior_analysis}

   WEB EVIDENCE (empty unless a search was run; never contradict it):
   {evidence}

   WHAT TO DO
   1. Turn each failed hard question into a short, imperative, unambiguous constraint and put them in `appended`. Prioritize count and position constraints — those are what the image model most reliably obeys.
   2. Judge your own factual certainty in `confidence` (0-1). This is about the WORLD KNOWLEDGE the constraints depend on (names, dates, formulas, quantities), not about your writing.
   3. If `confidence` < 0.7, set `need_search` to true and put 1-3 targeted `search_queries` in the output. Write real search queries about the missing facts — do NOT just echo the checklist question, it makes a poor query. Also include the word `need_search` in `reasoning`.
   4. If WEB EVIDENCE is present, ground the constraints in it and update `confidence`.

   HARD RULES
   - NEVER modify, shorten, or reorder the ORIGINAL PROMPT. `appended` is appended to it verbatim.
   - `appended` must start with: "STRICT LAYOUT/COUNT CONSTRAINTS (must all hold):"
   - One numbered line per constraint. Keep `appended` under 900 characters.
   - Never invent facts to fill a gap. If you do not know, lower `confidence` and ask for search.

   Return STRICT JSON only:
   {{"reasoning":"what knowledge these constraints depend on and how sure you are",
         "confidence":0.0-1.0,
         "need_search":true|false,
         "search_queries":["...","..."],
         "appended":"STRICT LAYOUT/COUNT CONSTRAINTS (must all hold):\n1. ..."}}
\end{promptbox}

\subsection{Retoucher}
\label{app:agent-prompts-retoucher}

\begin{promptbox}[Retoucher System Prompt]{CigarBlue}{CigarBlueBg}
   You are an infographic text repair expert with broad knowledge (chemistry, math, physics, biology, arts, history, astronomy).

   You are given: the failed checklist questions for a generated infographic, and the full OCR block list of that image. Your job is to decide, for each repairable block, what the CORRECT text should be.

   Priority of evidence, highest first:
   1. What the original prompt / checklist question explicitly requires.
   2. Established world knowledge (formulas, constants, dates, named entities).
   3. Inference of intent from the garbled string itself.

   Hard rules:
   - Never invent facts to fill a gap. If you cannot reach the required confidence, SKIP the block.
   - Only bind to block_index values that exist in the provided OCR list.
   - Keep replacement text about as long as the original; the renderer must fit it into the original bounding box, and a much longer string will be shrunk until unreadable.
   - Output strict JSON only, no commentary.
\end{promptbox}

\begin{promptbox}[Retoucher Prompt]{CigarBlue}{CigarBlueBg}
   Repair this infographic by answering failed checklist questions correctly, AND fixing leftover garbled blocks.

   === FAILED QUESTIONS (HARD prioritized, but fix EASY too) ===
   {failed_questions}

   === OCR TEXT BLOCKS (all blocks, including correct ones for context) ===
   {ocr_blocks}

   === CURRENT SCORES ===
   accuracy={accuracy:.2f}  n_fail_hard={n_fail_hard}  n_fail_easy={n_fail_easy}

   === GUIDANCE FROM ANALYZER (based on previous attempts on this image) ===
   {guidance}

   Instructions:
   1. For each failed question: give the correct answer, bind to the OCR block index to replace.
      {unbound_instruction}
   3. kind: 'formula' (math/chem -> LaTeX like $H_2O_2$, $x^{{2}}$, $\frac{{1}}{{2}}$) or 'text'|'label'|'title'|'value' (plain).
   4. confidence >= {min_confidence} required, or skip.
   5. Only bind to a block index that appears in the OCR list above. Never invent an index.
   6. The block indices above are specific to THIS list. Indices from any earlier round or from the guidance text may refer to a different OCR pass — always re-locate the block by its text content in the list above before using an index.
   7. If the guidance says a previous edit broke a previously-correct answer, do not repeat it: either leave that block alone or choose a replacement that satisfies both questions.

   Output JSON:
   {{"targets":[{{"qid":<int or null>,
                  "block_index":<int>,
                  "correct_text":"...",
                  "kind":"formula|text|label|title|value",
                  "latex_expr":"(if formula, same as correct_text)",
                  "is_hard":true|false,
                  "confidence":0-1}},...]}}
\end{promptbox}

\end{document}